\documentclass[11pt]{article}
\usepackage{acl2015}
\usepackage{times}
\usepackage{url}
\usepackage{latexsym}
\usepackage{verbatim}
\usepackage{tabularx}
\usepackage{rotating}
\usepackage{changepage}
\usepackage{graphicx}
\usepackage{amssymb}

\title{A Sensitivity Analysis of (and Practitioners' Guide to) Convolutional Neural Networks for Sentence Classification}
\author{Ye Zhang \\
  Dept. of Computer Science \\
  University of Texas at Austin \\
  {\tt yezhang@utexas.edu} \\\And
  Byron C. Wallace \\
  iSchool \\
  University of Texas at Austin \\
  {\tt byron.wallace@utexas.edu} \\}

\begin{document}
\maketitle
\begin{abstract}
Convolutional Neural Networks (CNNs) have recently achieved remarkably strong performance on the practically important task of sentence classification ~\cite{kim2014convolutional,kalchbrenner2014convolutional,johnson2014effective}. However, these models require practitioners to specify an exact model architecture and set accompanying hyperparameters, including the filter region size, regularization parameters, and so on. It is currently unknown how sensitive model performance is to changes in these configurations for the task of sentence classification. We thus conduct a sensitivity analysis of one-layer CNNs to explore the effect of architecture components on model performance; our aim is to distinguish between important and comparatively inconsequential design decisions for sentence classification. We focus on one-layer CNNs (to the exclusion of more complex models) due to their comparative simplicity and strong empirical performance, which makes it a modern standard baseline method akin to Support Vector Machine (SVMs) and logistic regression. We derive practical advice from our extensive empirical results for those interested in getting the most out of CNNs for sentence classification in real world settings.
\end{abstract}

%
%

%
%

%
%


\section{Introduction}
Convolutional Neural Networks (CNNs) have recently been shown to achieve impressive results on the practically important task of sentence categorization~\cite{kim2014convolutional,kalchbrenner2014convolutional,wang2semantic,goldberg2015primer,iyyer2015deep}. CNNs can capitalize on distributed representations of words by first converting the tokens comprising each sentence into a vector, forming a matrix to be used as input (e.g., see Fig. \ref{fig:CNN}).
The models need not be complex to realize strong results: Kim \shortcite{kim2014convolutional}, for example, proposed a simple one-layer CNN that achieved state-of-the-art (or comparable) results across several datasets. The very strong results achieved with this comparatively simple CNN architecture suggest that it may serve as a drop-in replacement for well-established baseline models, such as SVM~\cite{joachims1998text} or logistic regression. While more complex deep learning models for text classification will undoubtedly continue to be developed, those deploying such technologies in practice will likely be attracted to simpler variants, which afford fast training and prediction times.



Unfortunately, a downside to CNN-based models -- even simple ones -- is that they require practitioners to specify the exact model architecture to be used and to set the accompanying hyperparameters. To the uninitiated, making such decisions can seem like something of a black art because there are many free parameters in the model. This is especially true when compared to, e.g., SVM and logistic regression. Furthermore, in practice exploring the space of possible configurations for this model is extremely expensive, for two reasons: (1) training these models is relatively slow, even using GPUs. For example, on the SST-1 dataset \cite{socher2013recursive}, it takes about 1 hour to run 10-fold cross validation, using a similar configuration to that described in~\cite{kim2014convolutional}.\footnote{All experiments run with Theano on an NVIDIA K20 GPU.}
(2) The space of possible model architectures and hyperparameter settings is vast. Indeed, the simple CNN architecture we consider requires, at a minimum, specifying: input word vector representations; filter region size(s); the number of feature maps; the activation function(s); the pooling strategy; and regularization terms (dropout/$l2$). 

In practice, tuning all of these parameters is simply not feasible, especially because parameter estimation is computationally intensive. Emerging research has begun to explore hyperparameter optimization methods, including random search~\cite{bengio2012practical}, and Bayesian optimization~\cite{yogatama2015bayesian,bergstra2013making}.  However, these sophisticated search methods still require knowing which hyperparameters are worth exploring to begin with (and reasonable ranges for each). Furthermore, we believe it will be some time before Bayesian optimization methods are integrated into deployed, real-world systems.

In this work our aim is to identify empirically the settings that practitioners should expend effort tuning, and those that are either inconsequential with respect to performance or that seem to have a `best' setting independent of the specific dataset, and provide a reasonable range for each hyperparameter. We take inspiration from previous empirical analyses of neural models due to Coates et al. \shortcite{coates2011analysis} 
and Breuel~\shortcite{breuel2015effects}, which investigated factors in unsupervised feature learning and hyperparameter settings for Stochastic Gradient Descent (SGD), respectively. Here we report the results of a large number of experiments exploring different configurations of CNNs run over nine sentence classification datasets. Most previous work in this area reports only mean accuracies calculated via cross-validation. But there is substantial variance in the performance of CNNs, even on the \emph{same folds} and with model configuration held constant. Therefore, in our experiments we perform replications of cross-validation and report accuracy/Area Under Curve (AUC) score means and ranges over these.

For those interested in only the punchlines, we summarize our empirical findings and provide practical guidance based on these in Section \ref{section:conclusions}. 

\section{Background and Preliminaries}
\label{section:background}

%
Deep and neural learning methods are now well established in machine learning \cite{lecun2015deep,bengio2009learning}. They have been especially successful for image and speech processing tasks. More recently, such methods have begun to overtake traditional sparse, linear models for NLP \cite{goldberg2015primer,bengio2003neural,mikolov2013distributed,collobert2008unified,collobert2011natural,kalchbrenner2014convolutional,socher2013recursive}. 

%

Recently, word embeddings have been exploited for sentence classification using CNN architectures. Kalchbrenner \shortcite{kalchbrenner2014convolutional} proposed a CNN architecture with multiple convolution layers, positing latent, dense and low-dimensional word vectors (initialized to random values) as inputs. Kim \shortcite{kim2014convolutional} defined a one-layer CNN architecture that performed comparably. This model uses pre-trained word vectors as inputs, which may be treated as \emph{static} or \emph{non-static}. In the former approach, word vectors are treated as fixed inputs, while in the latter they are `tuned' for a specific task. Elsewhere, Johnson and Zhang \shortcite{johnson2014effective} proposed a similar model, but swapped in high dimensional `one-hot' vector representations of words as CNN inputs. 
Their focus was on classification of longer texts, rather than sentences (but of course the model can be used for sentence classification). 

The relative simplicity of Kim's architecture -- which is largely the same as that proposed by Johnson and Zhang \shortcite{johnson2014effective}, modulo the word vectors -- coupled with observed strong empirical performance makes this a strong contender to supplant existing text classification baselines such as SVM and logistic regression. But in practice one is faced with making several model architecture decisions and setting various hyperparameters. At present, very little empirical data is available to guide such decisions; addressing this gap is our aim here.

\vspace{-.25em}
\subsection{CNN Architecture}
\vspace{-.25em}
\label{CNN:architecture}

We begin with a tokenized sentence which we then convert to a \emph{sentence matrix}, the rows of which are word vector representations of each token. These might be, e.g., outputs from trained word2vec~\cite{mikolov2013distributed} or GloVe~\cite{pennington2014glove} models. We denote the dimensionality of the word vectors by $d$. If the length of a given sentence is $s$, then the dimensionality of the sentence matrix is $s \times d$.\footnote{We use the same zero-padding strategy as in~\cite{kim2014convolutional}.} Following Collobert and Weston \shortcite{collobert2008unified}, we can then effectively treat the sentence matrix as an `image', and perform convolution on it via linear \emph{filters}. In text applications there is inherent sequential structure to the data. Because rows represent discrete symbols (namely, words), it is reasonable to use filters with widths equal to the dimensionality of the word vectors (i.e., $d$). Thus we can simply vary the `height' of the filter, i.e., the number of adjacent rows considered jointly. We will refer to the height of the filter as the \emph{region size} of the filter.

Suppose that there is a filter parameterized by the weight matrix $\mathbf{w}$ with region size $h$; $\mathbf{w}$ will contain $h \cdot d$ parameters to be estimated. We denote the sentence matrix by $\mathbf{A}\in \mathbb{R}^{s\times d}$, and use $\mathbf{A}[i:j]$ to represent the sub-matrix of $\mathbf{A}$ from row $i$ to row $j$. The output sequence $\mathbf{o}\in\mathbb{R}^{s-h+1}$ of the convolution operator is obtained by repeatedly applying the filter on sub-matrices of $\mathbf{A}$:

\vspace{-.5em}


\begin{equation}
o_i = \mathbf{w}\cdot \mathbf{A}[i:i+h-1], 
\end{equation}

\noindent where $i=1 \ldots s-h+1$, and $\cdot$ is the dot product between the sub-matrix and the filter (a sum over element-wise multiplications). We add a bias term $b\in \mathbb{R}$ and an activation function $f$ to each $o_i$, inducing the \emph{feature map} $\mathbf{c} \in \mathbb{R}^{s-h+1}$ for this filter:

\vspace{-1em}
\begin{equation}
c_i = f(o_i+b).
\end{equation}
\vspace{-1.3em}

\noindent One may use multiple filters for the same region size to learn complementary features from the same regions. One may also specify multiple kinds of filters with different region sizes (i.e., `heights'). 

\begin{figure*}[!ht]
\centering
\includegraphics[width=142mm]{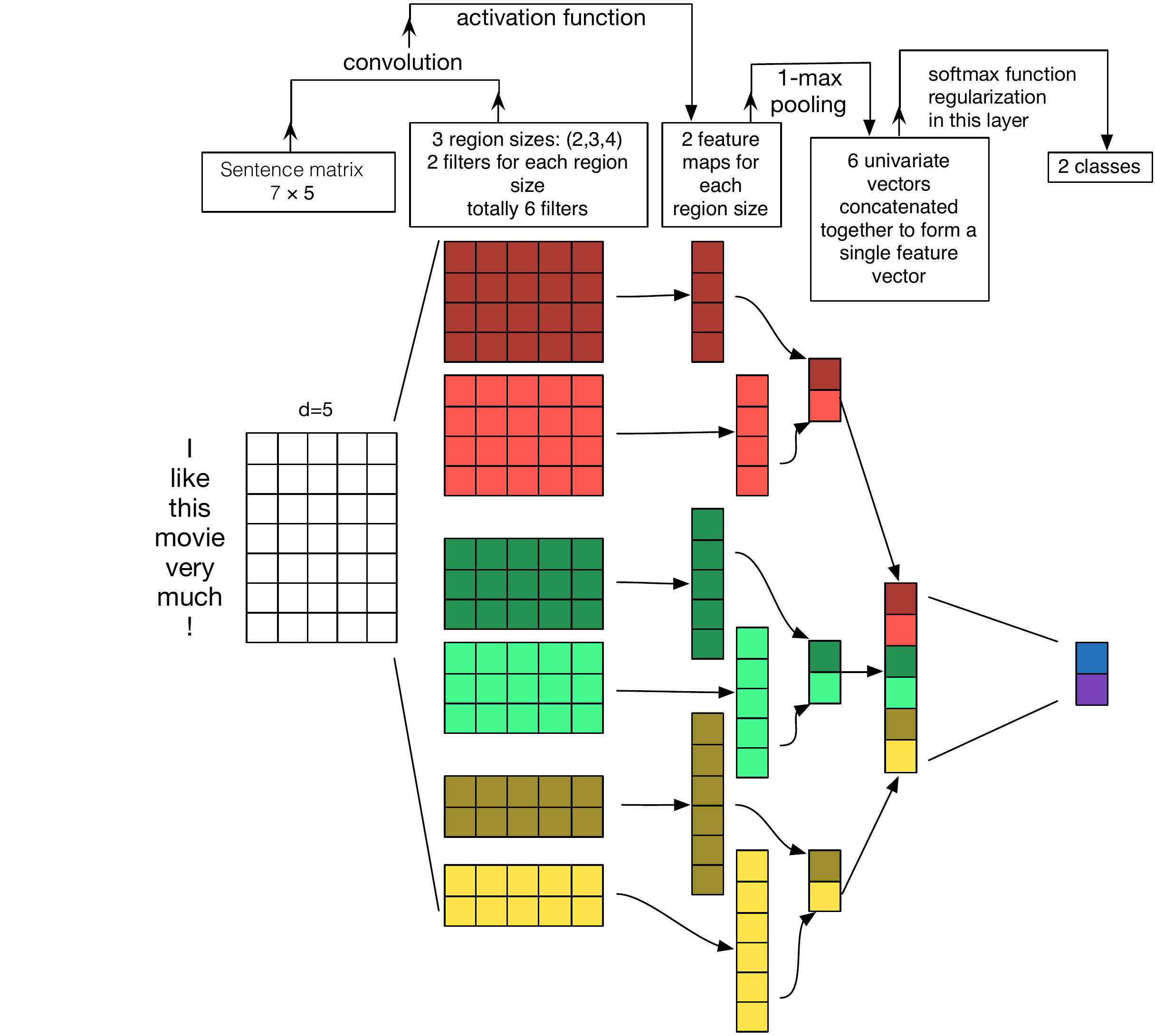}
\caption{Illustration of a CNN architecture for sentence classification. We depict three filter region sizes: 2, 3 and 4, each of which has 2 filters. Filters perform convolutions on the sentence matrix and generate (variable-length) feature maps; 1-max pooling is performed over each map, i.e., the largest number from each feature map is recorded. Thus a univariate feature vector is generated from all six maps, and these 6 features are concatenated to form a feature vector for the penultimate layer. The final softmax layer then receives this feature vector as input and uses it to classify the sentence; here we assume binary classification and hence depict two possible output states.}
\label{fig:CNN}
\vspace{-.1em}
\end{figure*}

The dimensionality of the feature map generated by each filter will vary as a function of the sentence length and the filter region size. A pooling function is thus applied to each feature map to induce a fixed-length vector. A common strategy is \emph{1-max pooling} ~\cite{boureau2010theoretical}, which extracts a scalar from each feature map. 
Together, the outputs generated from each filter map can be concatenated into a fixed-length, `top-level' feature vector, 
which is then fed through a softmax function to generate the final classification. At this softmax layer, one may apply `dropout'~\cite{hinton2012improving} as a means of regularization. This entails randomly setting values in the weight vector to 0. One may also impose an $l2$ norm constraint, i.e., linearly scale the $l2$ norm of the vector to a pre-specified threshold when it exceeds this. Fig. \ref{fig:CNN} provides a schematic illustrating the model architecture just described. 

A reasonable training objective to be minimized is the categorical cross-entropy loss. The parameters to be estimated include the weight vector(s) of the filter(s), the bias term in the activation function, and the weight vector of the softmax function. In the `non-static' approach, one also tunes the word vectors. Optimization is performed using SGD and back-propagation~\cite{rumelhart1988learning}. 

\section{Datasets}
\label{section:datasets}


We use nine sentence classification datasets in all; seven of which were also used by Kim~\shortcite{kim2014convolutional}. Briefly, these are summarized as follows. (1) \textbf{MR}: sentence polarity dataset from~\cite{Pang+Lee:05a}.
(2) \textbf{SST-1}: Stanford Sentiment Treebank~\cite{socher2013recursive}. To make input representations consistent across tasks, we only train and test on sentences, in contrast to the use in \cite{kim2014convolutional}, wherein models were trained on both phrases and sentences. (3) \textbf{SST-2}: Derived from SST-1, but pared to only two classes. We again only train and test models on sentences, excluding phrases. (4) \textbf{Subj}: Subjectivity dataset~\cite{Pang+Lee:05a}. (5) \textbf{TREC}: Question classification dataset~\cite{li2002learning}. (6) \textbf{CR}: Customer review dataset~\cite{hu2004mining}. (7) \textbf{MPQA}: Opinion polarity dataset~\cite{wiebe2005annotating}. 
Additionally, we use (8) \textbf{Opi}: Opinosis Dataset, which comprises sentences extracted from user reviews on a given topic, e.g. ``sound quality of ipod nano". There are 51 such topics and each topic contains approximately 100 sentences ~\cite{ganesan2010opinosis}. (9) \textbf{Irony}~\cite{wallace2014humans}: this contains 16,006 sentences from \emph{reddit} labeled as ironic (or not). The dataset is imbalanced (relatively few sentences are ironic). Thus before training, we under-sampled
negative instances to make classes sizes equal.\footnote{Empirically, under-sampling outperformed over-sampling in mitigating imbalance, see also Wallace~\shortcite{wallace2011class}.} For this dataset we report the Area Under Curve (AUC), rather than accuracy, because it is imbalanced.

\section{Baseline Models}
To provide a point of reference for the CNN results, we first report the performance achieved using SVM for sentence classification. As a baseline, we used a linear kernel SVM exploiting uni- and bi-gram features.\footnote{For this we used scikit-learn \cite{scikit-learn}.} We then used averaged word vectors (from Google word2vec\footnote{\url{https://code.google.com/p/word2vec/}} or GloVe\footnote{\url{http://nlp.stanford.edu/projects/glove/}}) calculated over the words comprising the sentence as features and used an RBF-kernel SVM as the classifier operating in this dense feature space. We also experimented with combining the uni-gram, bi-gram and word vector features with a linear kernel SVM. We kept only the most frequent 30k $n$-grams for all datasets, and tuned hyperparameters via nested cross-fold validation, optimizing for accuracy (AUC for Irony). For consistency, we used the same pre-processing steps for the data as described in previous work~\cite{kim2014convolutional}. We report means from 10-folds over all datasets in Table \ref{table:SVM}.\footnote{Note that parameter estimation for SVM via QP is deterministic, thus we do not replicate the cross validation here.} 
Notably, even naively incorporating word2vec embeddings into feature vectors usually improves results.

\begin{table}[h]
\footnotesize
\begin{center}
\begin{tabular}{c c c c}
\bf Dataset & \bf \bf {bowSVM} & \bf{wvSVM} & \bf {bowwvSVM}  \\ \hline
MR & 78.24  & 78.53 & 79.67 \\ 
SST-1 & 37.92  & 44.34 & 43.15\\  
SST-2 & 80.54 & 81.97 &83.30 \\ 
Subj & 89.13 & 90.94 & 91.74   \\ 
TREC &87.95 &83.61 & 87.33 \\ 
CR & 80.21 & 80.79& 81.31\\ 
MPQA &85.38 &89.27 & 89.70 \\ 
Opi &61.81 & 62.46 & 62.25 \\ 
Irony & 65.74 & 65.58 & 66.74  \\ 
\end{tabular}
\end{center}
\vspace{-1em}
\caption{Accuracy (AUC for Irony) achieved by SVM with different feature sets. {\bf bowSVM}: uni- and bi-gram features. {\bf wvSVM}: a naive word2vec-based representation, i.e., the average (300-dimensional) word vector for each sentence. {\bf bowwvSVM}: concatenates bow vectors with the average word2vec representations.}
\vspace{-1.5em}
\label{table:SVM}
\end{table}

\subsection{Baseline Configuration}
\label{section:baseline}
\vspace{-.2em} 

We first consider the performance of a baseline CNN configuration. Specifically, we start with the architectural decisions and hyperparameters used in previous work \cite{kim2014convolutional} and described in Table \ref{table:original}. To contextualize the variance in performance attributable to various architecture decisions and hyperparameter settings, it is critical to assess the variance due strictly to the parameter estimation procedure. Most prior work, unfortunately, has not reported such variance, despite a highly stochastic learning procedure. This variance is attributable to estimation via SGD, random dropout, and random weight parameter initialization. Holding all variables (including the folds) constant, we show that the mean performance calculated via 10-fold cross validation (CV) exhibits relatively high variance over repeated runs. We replicated CV experiments 100 times for each dataset, so that each replication was a 10-fold CV, wherein the folds were fixed. 
We recorded the average performance for each replication and report the mean, minimum and maximum average accuracy (or AUC) values observed over 100 replications of CV (that is, we report means and ranges of averages calculated over 10-fold CV). 
This provides a sense of the variance we might observe without any changes to the model. 
We did this for both static and non-static methods. For all experiments, we used the same preprocessing steps for the data as in~\cite{kim2014convolutional}. For SGD, we used the ADADELTA update rule ~\cite{zeiler2012adadelta}, and set the minibatch size to 50. We randomly selected 10\% of the training data as the validation set for early stopping. 

\begin{figure}[h]
\vspace{-.5em}
\includegraphics[width=0.5\textwidth]{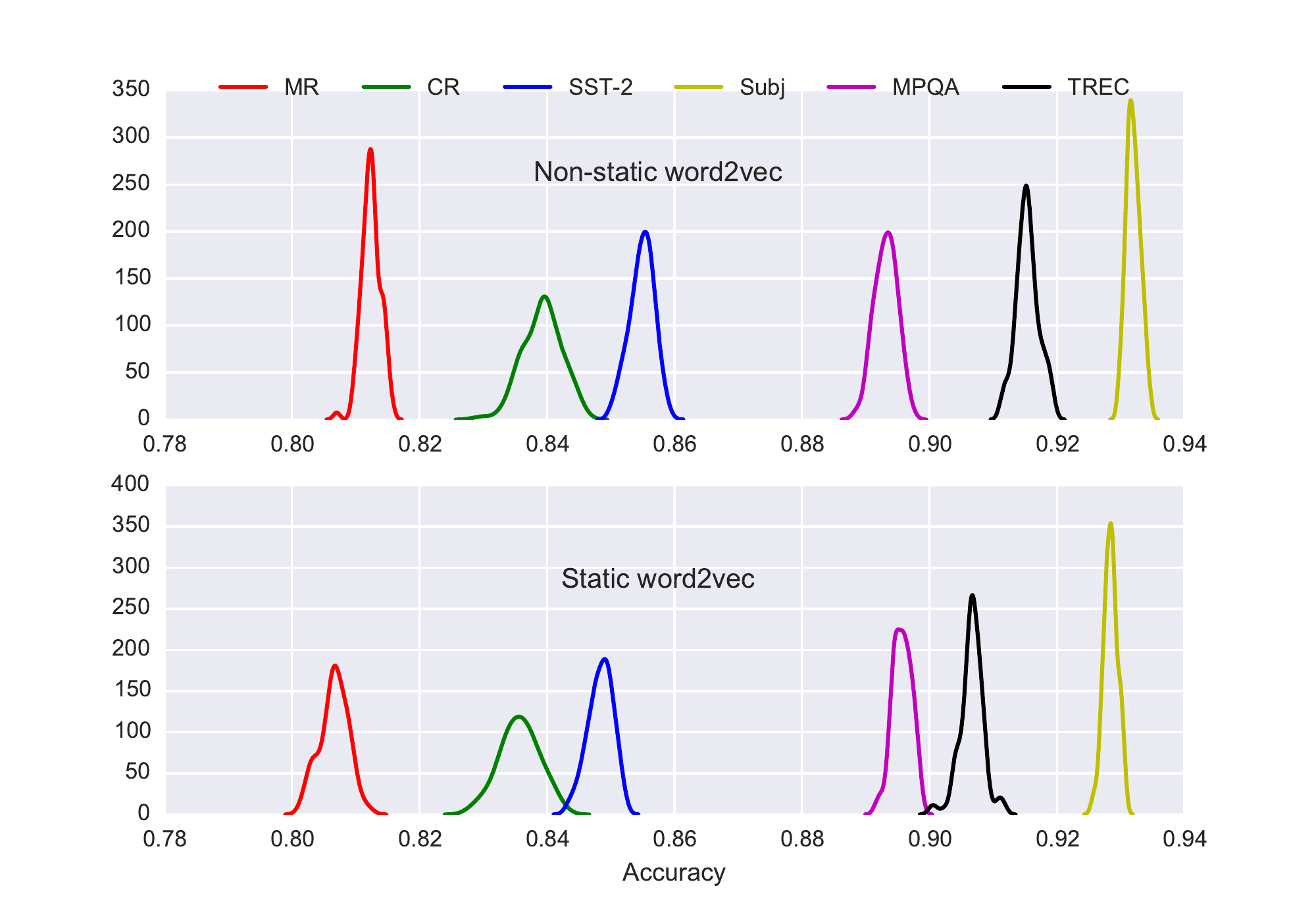}
\vspace{-2em}
\caption{Density curve of accuracy using static and non-static word2vec-CNN}
\label{fig:basic}
\vspace{-1em}
\end{figure}

Fig. \ref{fig:basic} provides density plots of the mean accuracy of 10-fold CV over the 100 replications for both methods on all datasets. For presentation clarity, in this figure we exclude the SST-1, Opi and Irony datasets, because performance was substantially lower on these (results can be found in the tables). Note that we pre-processed/split datasets differently than in some of the original work to ensure consistency for our present analysis; thus results may not be directly comparable to prior work. We emphasize that our aim here is not to improve on the state-of-the-art, but rather to explore the sensitivity of CNNs with respect to design decisions.


\begin{table}[h]
\footnotesize
\begin{center}
\begin{tabular}{c c}
\bf Description & \bf Values \\ \hline
input word vectors & Google word2vec  \\ 
filter region size & (3,4,5) \\ 
feature maps & 100 \\ 
activation function & ReLU \\ 
pooling & 1-max pooling \\ 
dropout rate & 0.5 \\ 
$l$2 norm constraint & 3  \\
\end{tabular}
\end{center}
\vspace{-.5em}
\caption{Baseline configuration. 
`feature maps' refers to the number of feature maps for each filter region size. `ReLU' refers to \emph{rectified linear unit} ~\protect\cite{maas2013rectifier}, a commonly used activation function in CNNs.}
\label{table:original}
\vspace{-.75em}
\end{table}

\begin{table*}[htbp]
\footnotesize
\centering
\begin{tabular}{c c c c}
\bf Dataset & \bf {Non-static word2vec-CNN} &  \bf{Non-static GloVe-CNN} &\bf{Non-static GloVe+word2vec CNN}\\ \hline
 MR & 81.24 (80.69, 81.56)&81.03 (80.68,81.48) &81.02 (80.75,81.32)\\ 
 SST-1 & 47.08 (46.42,48.01)&45.65 (45.09,45.94)&45.98 (45.49,46.65) \\ 
 SST-2 & 85.49 (85.03, 85.90)&85.22 (85.04,85.48)& 85.45 (85.03,85.82)\\ 
Subj & 93.20 (92.97, 93.45)& 93.64 (93.51,93.77) & 93.66 (93.39,93.87)\\ 
 TREC & 91.54 (91.15, 91.92)&90.38 (90.19,90.59)&91.37 (91.13,91.62) \\ 
 CR & 83.92 (82.95, 84.56) &84.33 (84.00,84.67)&84.65 (84.21,84.96) \\ 
 MPQA & 89.32 (88.84, 89.73)&89.57 (89.31,89.78)&89.55 (89.22,89.88)\\ 
 Opi & 64.93 (64.23,65.58) &65.68 (65.29,66.19)&65.65 (65.15,65.98) \\ 
 Irony &67.07 (65.60,69.00)&67.20 (66.45,67.96)&67.11 (66.66,68.50)\\ 
\end{tabular}
\caption{Performance using non-static word2vec-CNN, non-static GloVe-CNN, and non-static GloVe+word2vec CNN, respectively. Each cell reports the mean (min, max) of summary performance measures calculated over multiple runs of 10-fold cross-validation. We will use this format for all tables involving replications}
\label{table:basic}
\vspace{-.75em}
\end{table*}

Having established a baseline performance for CNNs, we now consider the effect of different architecture decisions and hyperparameter settings. To this end, we hold all other settings constant (as per Table \ref{table:original}) and vary only the component of interest.
For every configuration that we consider, we replicate the experiment 10 times, where each replication again constitutes a run of 10-fold CV.\footnote{Running 100 replications for every configuration that we consider was not computationally feasible.} We again report average CV means and associated ranges achieved over the replicated CV runs. We performed experiments using both `static' and `non-static' word vectors. The latter uniformly outperformed the former, and so here we report results only for the `non-static' variant. 

\subsection{Effect of input word vectors}
\vspace{-.5em}
\label{section:effect-of-wv}

A nice property of sentence classification models that start with distributed representations of words as inputs is the flexibility such architectures afford to swap in different pre-trained word vectors during model initialization. Therefore, we first explore the sensitivity of CNNs for sentence classification with respect to the input representations used. Specifically, we replaced word2vec with GloVe representations. Google word2vec uses a local context window model trained on 100 billion words from Google News~\cite{mikolov2013distributed}, while GloVe is a model based on global word-word co-occurrence statistics~\cite{pennington2014glove}. We used a GloVe model trained on a corpus of 840 billion tokens of web data. For both word2vec and GloVe we induce 300-dimensional word vectors. 
We report results achieved using GloVe representations in Table \ref{table:basic}. Here we only report non-static GloVe results (which again uniformely outperformed the static variant).

We also experimented with concatenating word2vec and GloVe representations, thus creating 600-dimensional word vectors to be used as input to the CNN. Pre-trained vectors may not always be available for specific words  (either in word2vec or GloVe, or both); in such cases, we randomly initialized the corresponding sub-vectors.
Results are reported in the final column of Table \ref{table:basic}. 

The relative performance achieved using GloVe versus word2vec 
depends on the dataset, and, unfortunately, simply concatenating these representations does necessarily seem helpful. Practically, our results suggest experimenting with different pre-trained word vectors for new tasks.

We also experimented with using long, sparse one-hot vectors as input word representations, in the spirit of Johnson and Zhang \shortcite{johnson2014effective}. In this strategy, each word is encoded as a one-hot vector, with dimensionality equal to the vocabulary size. Though this representation combined with one-layer CNN achieves good results on document classification, it is still unknown whether this is useful for sentence classification. 
We keep the other settings the same as in the basic configuration, and the one-hot vector is fixed during training. Compared to using embeddings as input to the CNN, we found the one-hot approach to perform poorly for sentence classification tasks. We believe that one-hot CNN may not be suitable for sentence classification when one has a small to modestly sized training dataset, likely due to sparsity: the sentences are perhaps too brief to provide enough information for this high-dimensional encoding. Alternative one-hot architectures might be more appropriate for this scenario. For example, Johnson and Zhang \cite{johnson2015semi} propose a semi-supervised CNN variant which first learns embeddings of small text regions from unlabeled data, and then integrates them into a supervised CNN. We emphasize that if training data is plentiful, learning embeddings from scratch may indeed be best.

\subsection{Effect of filter region size}
\label{section:effect-filter-region-size}
\vspace{-.5em}

\begin{table}[h]
\footnotesize
\begin{center}
\begin{tabular}{c c}
\bf Region size & \bf MR  \\ \hline 
 1 & 77.85 (77.47,77.97)\\ 
 3 & 80.48 (80.26,80.65)\\ 
 5 & 81.13 (80.96,81.32) \\ 
 7 & \bf{81.65 (81.45,81.85)} \\
 10 & 81.43 (81.28,81.75) \\ 
 15 & 81.26 (81.01,81.43) \\ 
 20 & 81.06 (80.87,81.30) \\ 
 25 & 80.91 (80.73,81.10) \\ 
 30 & 80.91 (80.72,81.05)\\ 
\end{tabular}
\end{center}
\vspace{-.5em}
\caption{Effect of single filter region size. Due to space constraints, we report results for only one dataset here, but these are generally illustrative.}
\label{table:filter_size}
\vspace{-1em}
\end{table}

\begin{figure}
\centering
\includegraphics[width=0.425\textwidth]{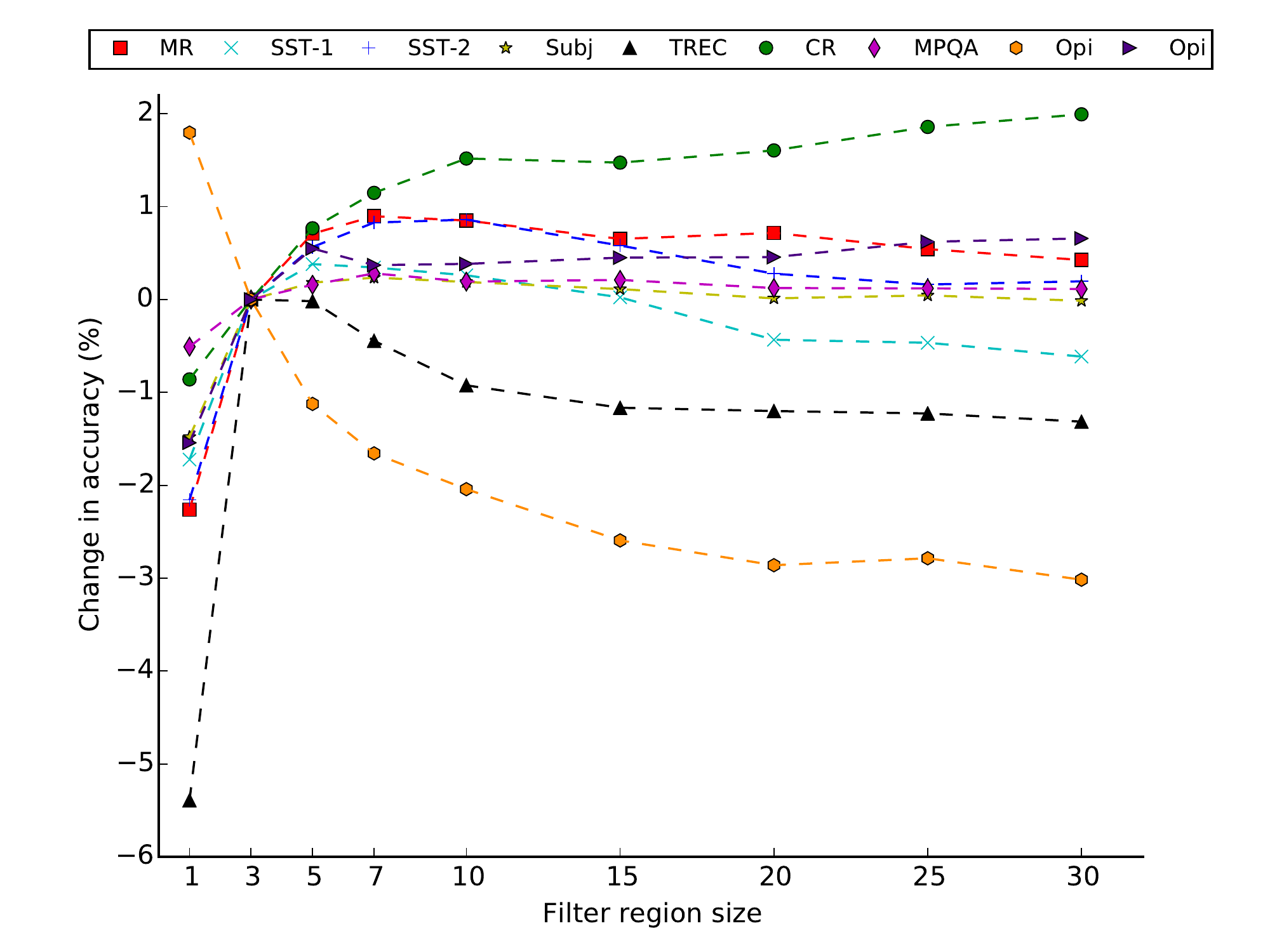}
\vspace{-.75em}
\caption{Effect of the region size (using only one).}
\label{fig:nonstatic_filter}
\vspace{-.5em}
\end{figure}

We first explore the effect of filter region size when using only one region size, and we set the number of feature maps for this region size to 100 (as in the baseline configuration). We consider region sizes of 1, 3, 5, 7, 10, 15, 20, 25 and 30, and record the means and ranges over 10 replications of 10-fold CV for each. We report results in Table \ref{table:filter_size} and Fig. \ref{fig:nonstatic_filter}. Because we are only interested in the trend of the accuracy as we alter the region size (rather than the absolute performance on each task), we show only the percent change in accuracy (AUC for Irony) from an arbitrary baseline point (here, a region size of 3). 

\begin{figure}
\centering
\includegraphics[width=0.425\textwidth]{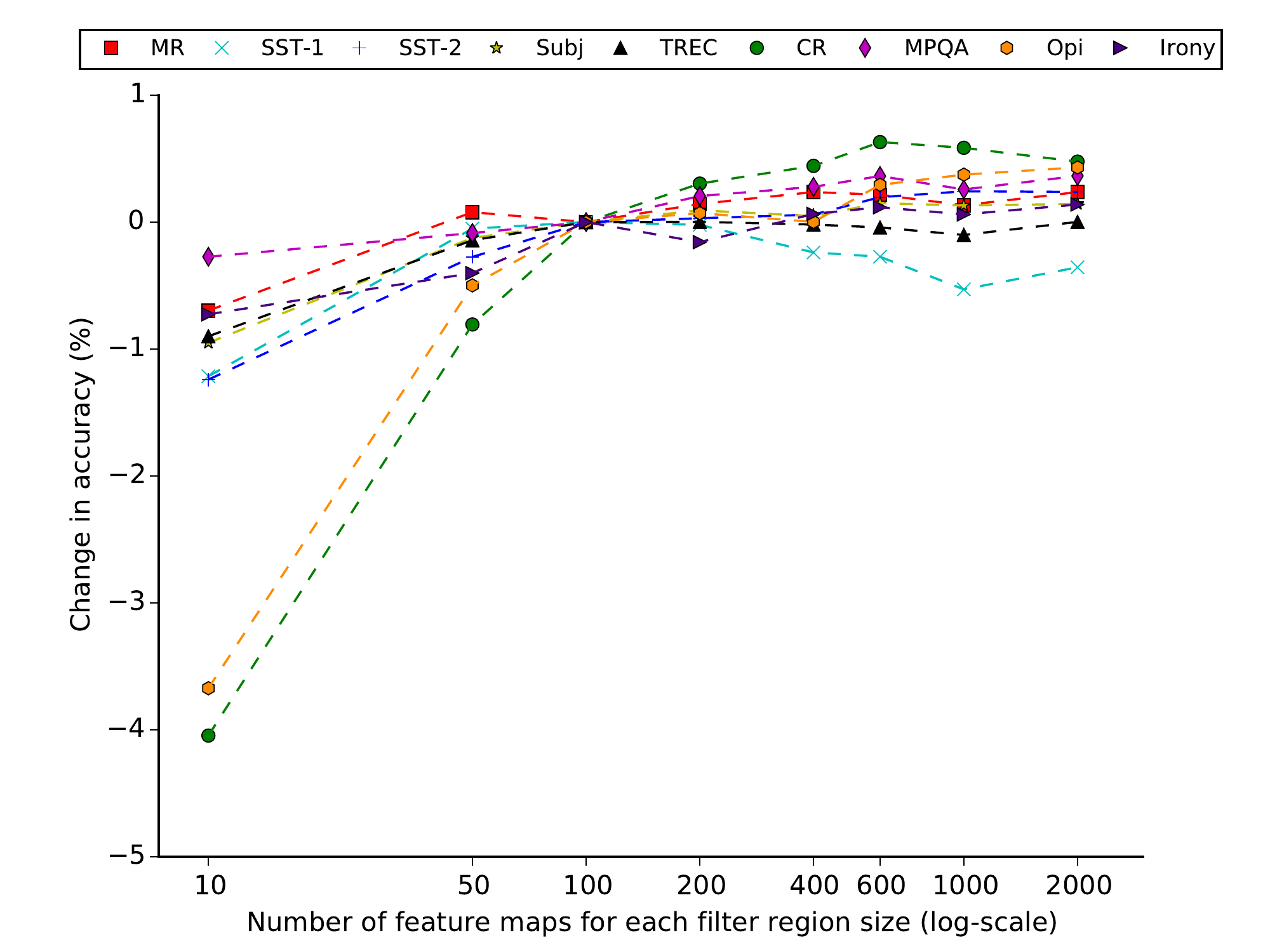}
\vspace{-.75em}
\caption{Effect of the number of feature maps.}
\vspace{-.75em}
\label{fig:nonstatic_feature_map}
\end{figure}

From the results, one can see that each dataset has its own optimal filter region size. Practically, this suggests performing a coarse grid search over a range of region sizes; the figure here suggests that a reasonable range for sentence classification might be from 1 to 10. However, for datasets comprising longer sentences, such as CR (maximum sentence length is 105, whereas it ranges from 36-56 on the other sentiment datasets used here), the optimal region size may be larger. 


We also explored the effect of combining different filter region sizes, while keeping the number of feature maps for each region size fixed at 100. We found that combining several filters with region sizes close to the optimal single region size can improve performance, but adding region sizes far from the optimal range may hurt performance. For example, when using a single filter size, one can observe that the optimal single region size for the MR dataset is 7. We therefore combined several different filter region sizes close to this optimal range, and compared this to approaches that use region sizes outside of this range. From Table \ref{table:multiple_filter_size}, one can see that using (5,6,7),and (7,8,9) and (6,7,8,9) -- sets near the best single region size -- produce the best results. The difference is especially pronounced when comparing to the baseline setting of (3,4,5). Note that even only using a single good filter region size (here, 7) results in better performance than combining different sizes (3,4,5). The best performing strategy is to simply use many feature maps (here, 400) all with region size equal to 7, i.e., the single best region size. 
However, we note that in some cases (e.g., for the TREC dataset), using multiple different, but near-optimal, region sizes performs best. 
\begin{table}
\centering
\footnotesize
\begin{tabular}{c c}
  \hline
\bf {Multiple region size} & \bf {Accuracy (\%)} \\ \hline
(7) & 81.65 (81.45,81.85) \\ 
(3,4,5) &81.24 (80.69, 81.56) \\ 
(4,5,6) & 81.28 (81.07,81.56) \\ 
(5,6,7) & 81.57 (81.31,81.80) \\ 
(7,8,9) & 81.69 (81.27,81.93)\\ 
(10,11,12) & 81.52 (81.27,81.87) \\ 
(11,12,13) & 81.53 (81.35,81.76) \\ 
(3,4,5,6) & 81.43 (81.10,81.61) \\ 
(6,7,8,9) & 81.62 (81.38,81.72) \\ 
(7,7,7) &  81.63 (81.33,82.08) \\ 
\bf (7,7,7,7) &\bf 81.73 (81.33,81.94) \\  
\end{tabular}
\caption{Effect of filter region size with several region sizes on the MR dataset.}
\label{table:multiple_filter_size}
\vspace{-1em}
\end{table}

We provide another illustrative empirical result using several region sizes on the TREC dataset in Table \ref{table:multiple_filter_size_TREC}. From the performance of single region size, we see that the best single filter region sizes for TREC are 3 and 5, so we explore the region size around these values, and compare this to using multiple region sizes far away from these `optimal' values. 

\begin{table}[!h]
\centering
\begin{tabular}{c c}
\bf {Multiple region size} & \bf {Accuracy (\%)} \\ \hline
(3) & 91.21 (90.88,91.52) \\ 
(5) & 91.20 (90.96,91.43) \\ 
(2,3,4) & 91.48 (90.96,91.70) \\ 
(3,4,5) & 91.56 (91.24,91.81) \\ 
(4,5,6) & 91.48 (91.17,91.68) \\ 
(7,8,9) & 90.79 (90.57,91.26) \\ 
(14,15,16) & 90.23 (89.81,90.51) \\ 
\bf (2,3,4,5) & \bf 91.57 (91.25,91.94) \\ 
(3,3,3) & 91.42 (91.11,91.65) \\ 
(3,3,3,3) & 91.32 (90.53,91.55) \\ 
\end{tabular}
\caption{Effect of filter region size with several region sizes using non-static word2vec-CNN on TREC dataset}
\label{table:multiple_filter_size_TREC}
\end{table}

Here we see that (3,3,3) and (3,3,3,3) perform worse than (2,3,4) and (3,4,5). However, the result still shows that a combination of region sizes near the optimal single best region size outperforms using multiple region sizes far from the optimal single region size. Furthermore, we again see that a single good region size (3) outperforms combining several suboptimal region sizes: (7,8,9) and (14,15,16). 

In light of these observations, we believe it advisable to first perform a coarse line-search over a single filter region size to find the `best' size for the dataset under consideration, and then explore the combination of several region sizes nearby this single best size, including combining both different region sizes and copies of the optimal sizes.

\subsection{Effect of number of feature maps for each filter region size}

We again hold other configurations constant, and thus have three filter region sizes: 3, 4 and 5. We change only the number of feature maps for each of these relative to the baseline of 100; we consider values $\in$ \{10, 50, 100, 200, 400, 600, 1000, 2000\}. We report results in Fig. \ref{fig:nonstatic_feature_map}.

The `best' number of feature maps for each filter region size depends on the dataset. However, it would seem that increasing the number of maps beyond 600 yields at best very marginal returns, and often hurts performance (likely due to overfitting). Another salient practical point is that it takes a longer time to train the model when the number of feature maps is increased. 
In practice, the evidence here suggests perhaps searching over a range of 100 to 600. Note that this range is only provided as a possible standard trick when one is faced with a new similar sentence classification problem; it is of course possible that in some cases more than 600 feature maps will be beneficial, but the evidence here suggests expending the effort to explore this is probably not worth it. In practice, one should consider whether the best observed value falls near the border of the range searched over; if so, it is probably worth exploring beyond that border, as suggested in~\cite{bengio2012practical}.

\subsection{Effect of activation function}
\label{section:activation-func} 

We consider seven different activation functions in the convolution layer, including: ReLU (as per the baseline configuration), hyperbolic tangent (tanh), Sigmoid function~\cite{maas2013rectifier}, SoftPlus function~\cite{dugas2001incorporating}, Cube function~\cite{chen2014fast}, and tanh cube function~\cite{pei2015effective}. We use `Iden' to denote the identity function, which means not using any activation function. We report results achieved using different activation functions in non-static CNN in Table \ref{table:activation}.

\begin{table*}[t]
\footnotesize
\begin{adjustwidth}{}{}
\centering
\begin{tabular}{c c c c c}
\bf Dataset & \bf tanh &\bf Softplus &\bf Iden &\bf ReLU \\ \hline
MR  & 81.28 (81.07, 81.52) & 80.58 (80.17, 81.12) & \bf {81.30 (81.09, 81.52)} & 81.16 (80.81, 83.38)\\ 

SST-1 & 47.02 (46.31, 47.73) & 46.95 (46.43, 47.45) & 46.73 (46.24,47.18) & \bf {47.13 (46.39, 47.56)} \\ 

SST-2 &\bf {85.43 (85.10, 85.85)} & 84.61 (84.19, 84.94) & 85.26 (85.11, 85.45) & 85.31 (85.93, 85.66) \\ 

Subj &  \bf {93.15 (92.93, 93.34)} & 92.43 (92.21, 92.61) & 93.11 (92.92, 93.22) & 93.13 (92.93, 93.23) \\ 

TREC & 91.18 (90.91, 91.47)& 91.05 (90.82, 91.29) & 91.11 (90.82, 91.34) & \bf {91.54 (91.17, 91.84)} \\ 

CR  & 84.28 (83.90, 85.11) & 83.67 (83.16, 84.26)&\bf{84.55 (84.21, 84.69)}& 83.83 (83.18, 84.21) \\ 

MPQA & 89.48 (89.16, 89.84) & \bf {89.62 (89.45, 89.77)} & 89.57 (89.31, 89.88) & 89.35 (88.88, 89.58) \\ 

Opi & \bf {65.69 (65.16,66.40)} & 64.77 (64.25,65.28) & 65.32 (64.78,66.09) & 65.02 (64.53,65.45)  \\

Irony & \bf{ 67.62 (67.18,68.27)} & 66.20 (65.38,67.20) & 66.77 (65.90,67.47) & 66.46 (65.99,67.17) \\ 
\end{tabular}
\caption{Performance of different activation functions}
\label{table:activation}
\end{adjustwidth}
\vspace{-.5em}
\end{table*}

For 8 out of 9 datasets, the best activation function is one of Iden, ReLU and tanh. The SoftPlus function outperformedd these on only one dataset (MPQA). 
Sigmoid, Cube, and tanh cube all consistently performed worse than alternative activation functions. Thus we do not report results for these here. The performance of the tanh function may be due to its zero centering property (compared to Sigmoid). ReLU has the merits of a non-saturating form compared to Sigmoid, and it
has been observed to accelerate the convergence of SGD~\cite{krizhevsky2012imagenet}. 
One interesting result is that not applying any activation function (Iden) sometimes helps. This indicates that on some datasets, a linear transformation is enough to capture the correlation between the word embedding and the output label. However, if there are multiple hidden layers, Iden may be less suitable than non-linear activation functions. Practically, with respect to the choice of the activation function in one-layer CNNs, our results suggest experimenting with ReLU and tanh, and perhaps also Iden.

\subsection{Effect of pooling strategy}

We next investigated the effect of the pooling strategy and the pooling region size. We fixed the filter region sizes and the number of feature maps as in the baseline configuration, thus changing only the pooling strategy or pooling region size.


In the baseline configuration, we performed 1-max pooling globally over feature maps, inducing a feature vector of length 1 for each filter. However, pooling may also be performed over small equal sized local regions rather than over the entire feature map \cite{boureau2011ask}. Each small local region on the feature map will generate a single number from pooling, and these numbers can be concatenated to form a feature vector for one feature map. The following step is the same as 1-max pooling: we concatenate all the feature vectors together to form a single feature vector for the classification layer. We experimented with local region sizes of 3, 10, 20, and 30, and found that 1-max pooling outperformed all local max pooling configurations. This result held across all datasets.

We also considered a $k$-max pooling strategy similar to \cite{kalchbrenner2014convolutional}, in which the maximum $k$ values are extracted from the entire feature map, and the relative order of these values is preserved. We explored the $k\in\{5,10,15,20\}$, and again found 1-max pooling fared best, consistently outperforming $k$-max pooling. 

Next, we considered taking an average, rather than the max, over regions \cite{boureau2010learning}. We held the rest of architecture constant. We experimented with local average pooling region sizes \{3, 10, 20, 30\}. We found that average pooling uniformly performed (much) worse than max pooling, at least on the CR and TREC datasets. 
Due to the substantially worse performance and very slow running time observed under average pooling, we did not complete experiments on all datasets.

Our analysis of pooling strategies shows that 1-max pooling consistently performs better than alternative strategies for the task of sentence classification. This may be because the location of predictive contexts does not matter, and certain $n$-grams in the sentence can be more predictive on their own than the entire sentence considered jointly. 


\subsection{Effect of regularization}
\label{section:regularization}

\begin{figure}
\centering
\includegraphics[width=0.425\textwidth]{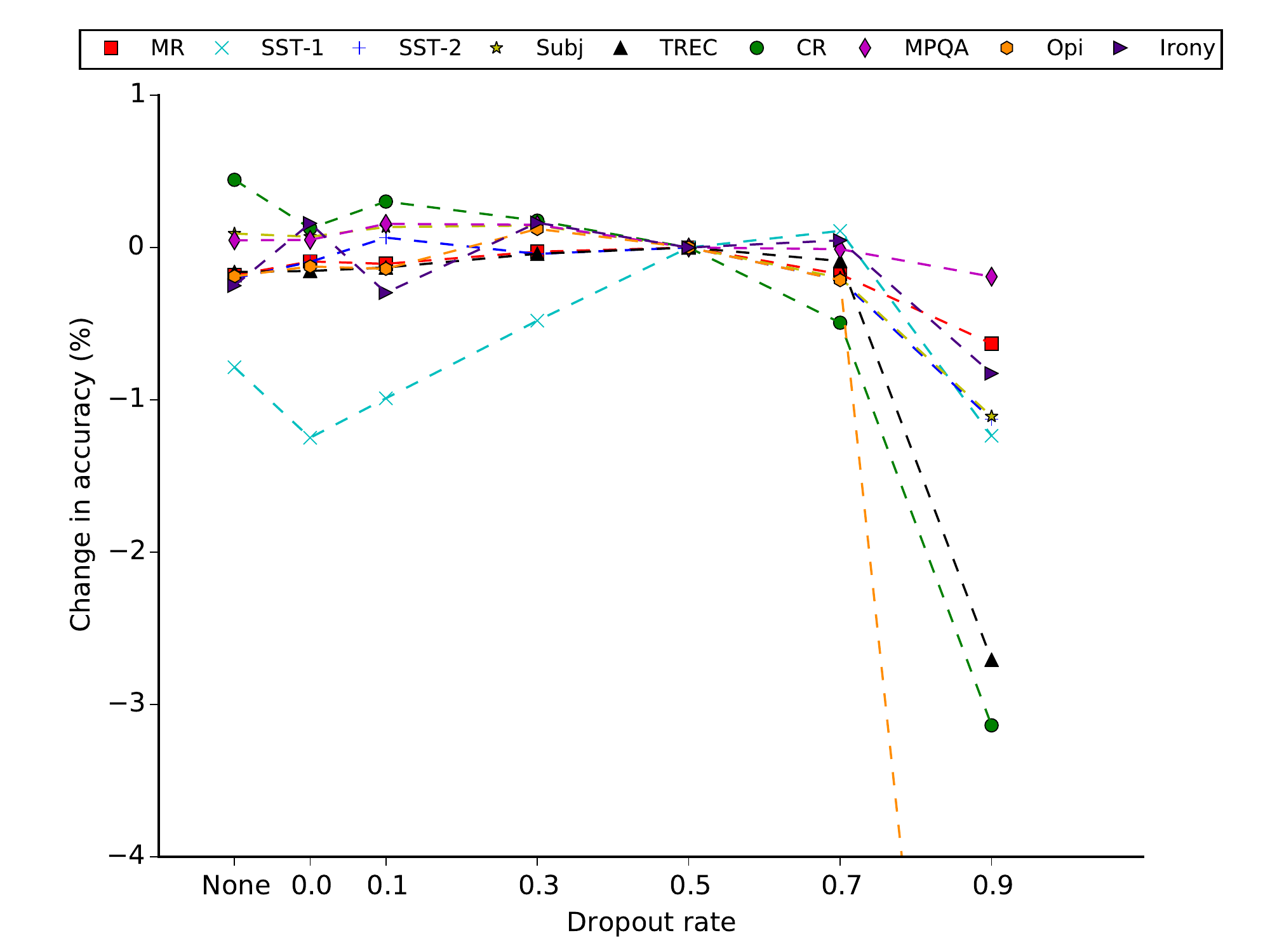}
\caption{Effect of dropout rate. The accuracy when the dropout rate is 0.9 on the Opi dataset is about 10\% worse than baseline, and thus is not visible on the figure at this point.}
\vspace{-1em}
\label{fig:nonstatic_dropout}
\end{figure}

Two common regularization strategies for CNNs are dropout and $l2$ norm constraints; we explore the effect of these here. `Dropout' is applied to the input to the penultimate layer. We experimented with varying the dropout rate from 0.0 to 0.9, fixing the $l2$ norm constraint to 3, as per the baseline configuration. The results for non-static CNN are shown in in Fig. \ref{fig:nonstatic_dropout}, with 0.5 designated as the baseline. We also report the accuracy achieved when we remove both dropout and the $l2$ norm constraint (i.e., when no regularization is performed), denoted by `None'. 

\begin{figure}[h]
\centering
\includegraphics[width=0.425\textwidth]{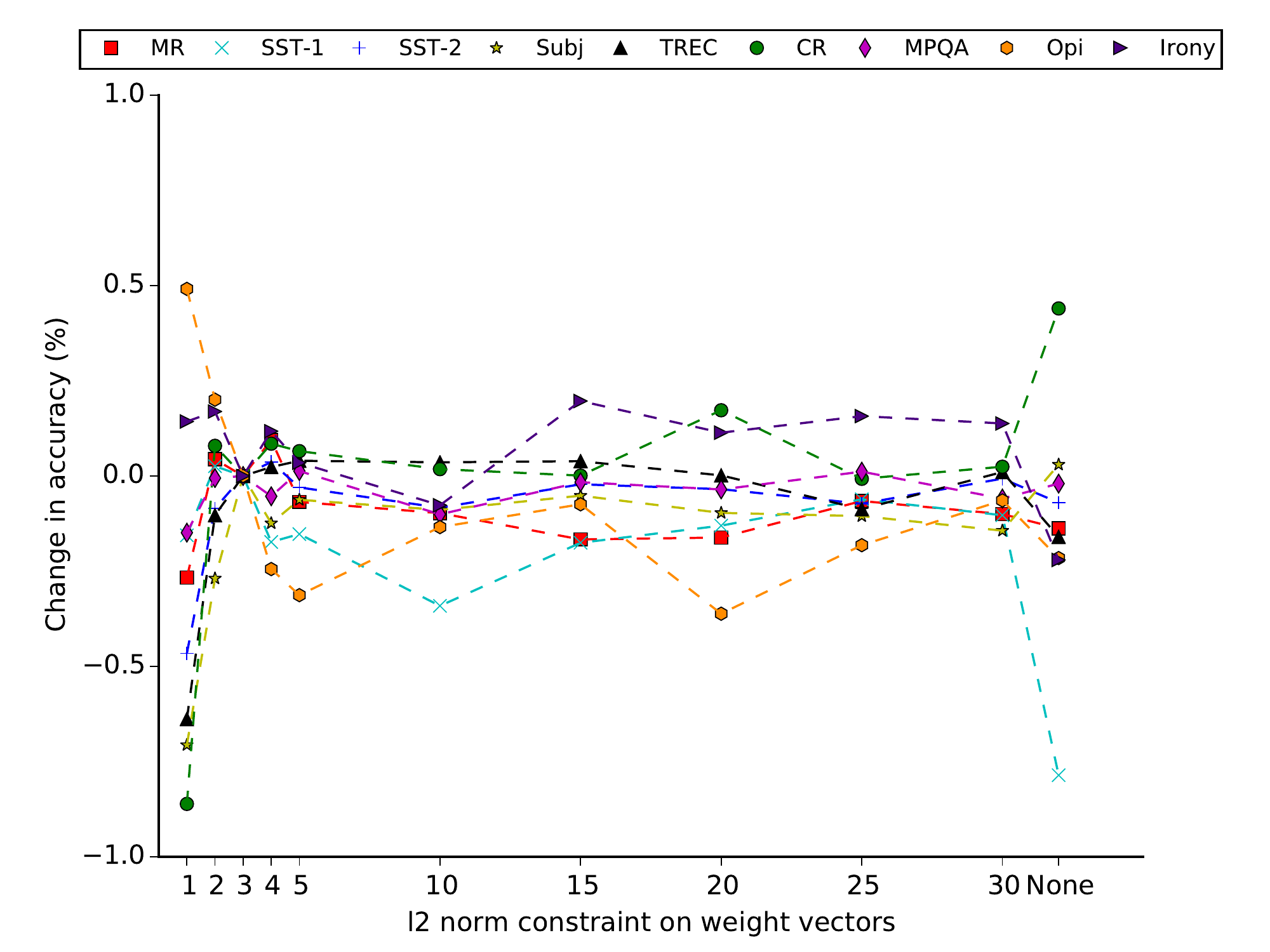}
\caption{Effect of the $l2$ norm constraint on weight vectors.}
\label{fig:l2}
\end{figure}

Separately, we considered the effect of the $l2$ norm imposed on the weight vectors that parametrize the softmax function. Recall that the $l2$ norm of a weight vector is linearly scaled to a constraint $c$ when it exceeds this threshold, so a smaller $c$ implies stronger regularization. (Like dropout, this strategy is applied only to the penultimate layer.) We show the relative effect of varying $c$ on non-static CNN in Figure \ref{fig:l2}, where we have fixed the dropout rate to 0.5; 3 is the baseline here (again, arbitrarily).

From Figures \ref{fig:nonstatic_dropout} and \ref{fig:l2}, one can see that non-zero dropout rates can help (though very little) at some points from 0.1 to 0.5, depending on datasets. But imposing an $l2$ norm constraint generally does not improve performance much (except on Opi), and even adversely effects performance on at least one dataset (CR). 

\begin{figure}[h]
\centering
\includegraphics[width=0.425\textwidth]{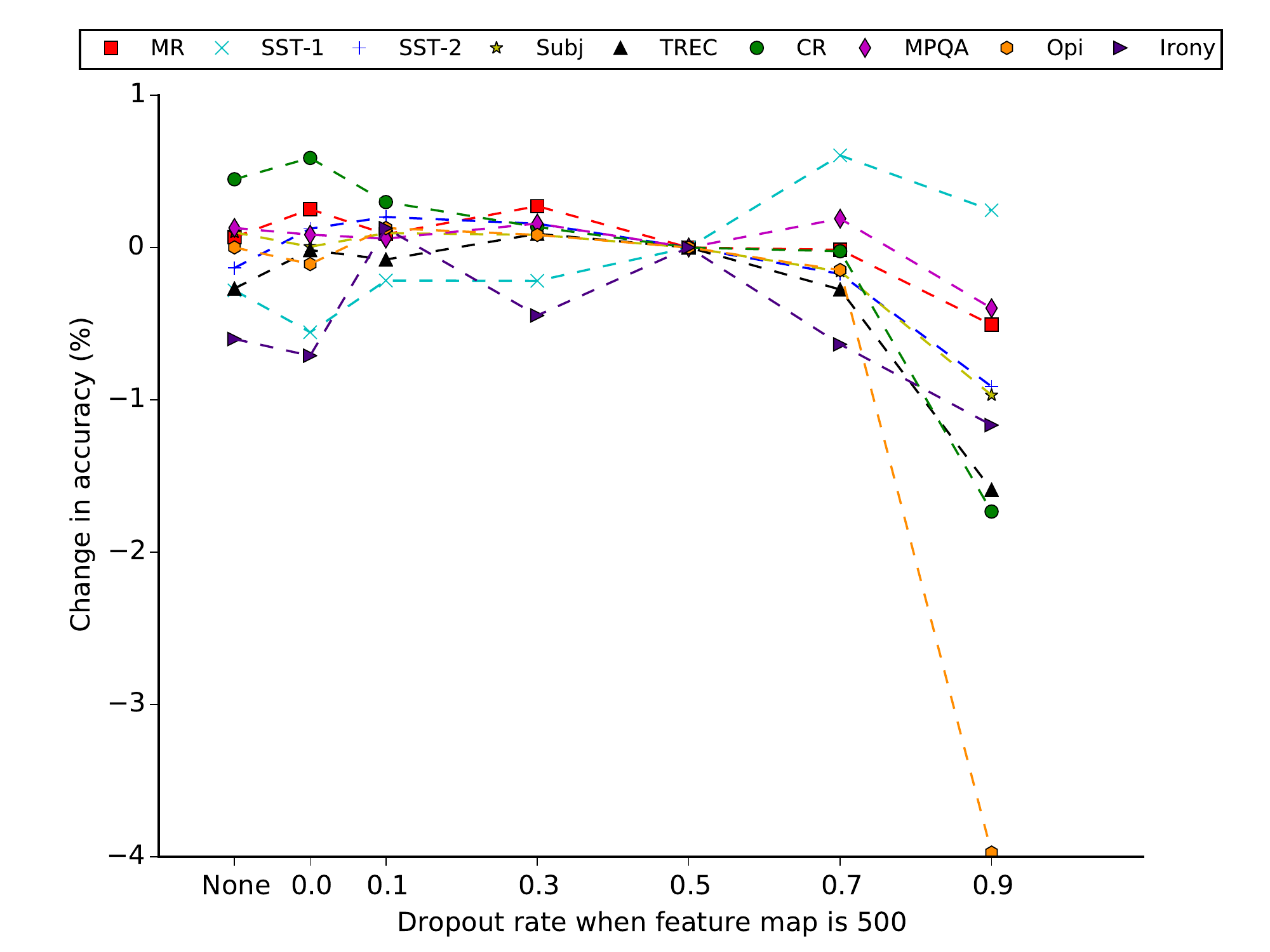}
\caption{Effect of dropout rate when using 500 feature maps.}
\label{fig:dropout_large}
\end{figure}

We then also explored dropout rate effect when increasing the number of feature maps. We increase the number of feature maps for each filter size from 100 to 500, and set max $l2$ norm constraint as 3. The effect of dropout rate is shown in Fig. \ref{fig:dropout_large}. 
We see that the effect of dropout rate is almost the same as when the number of feature maps is 100, and it does not help much. But we observe that for the dataset SST-1, dropout rate actually helps when it is 0.7. Referring to Fig. \ref{fig:nonstatic_feature_map}, we can see that when the number of feature maps is larger than 100, it hurts the performance possibly due to overfitting, so it is reasonable that in this case dropout would mitigate this effect.

\begin{figure}[h]
\centering
\includegraphics[width=0.425\textwidth]{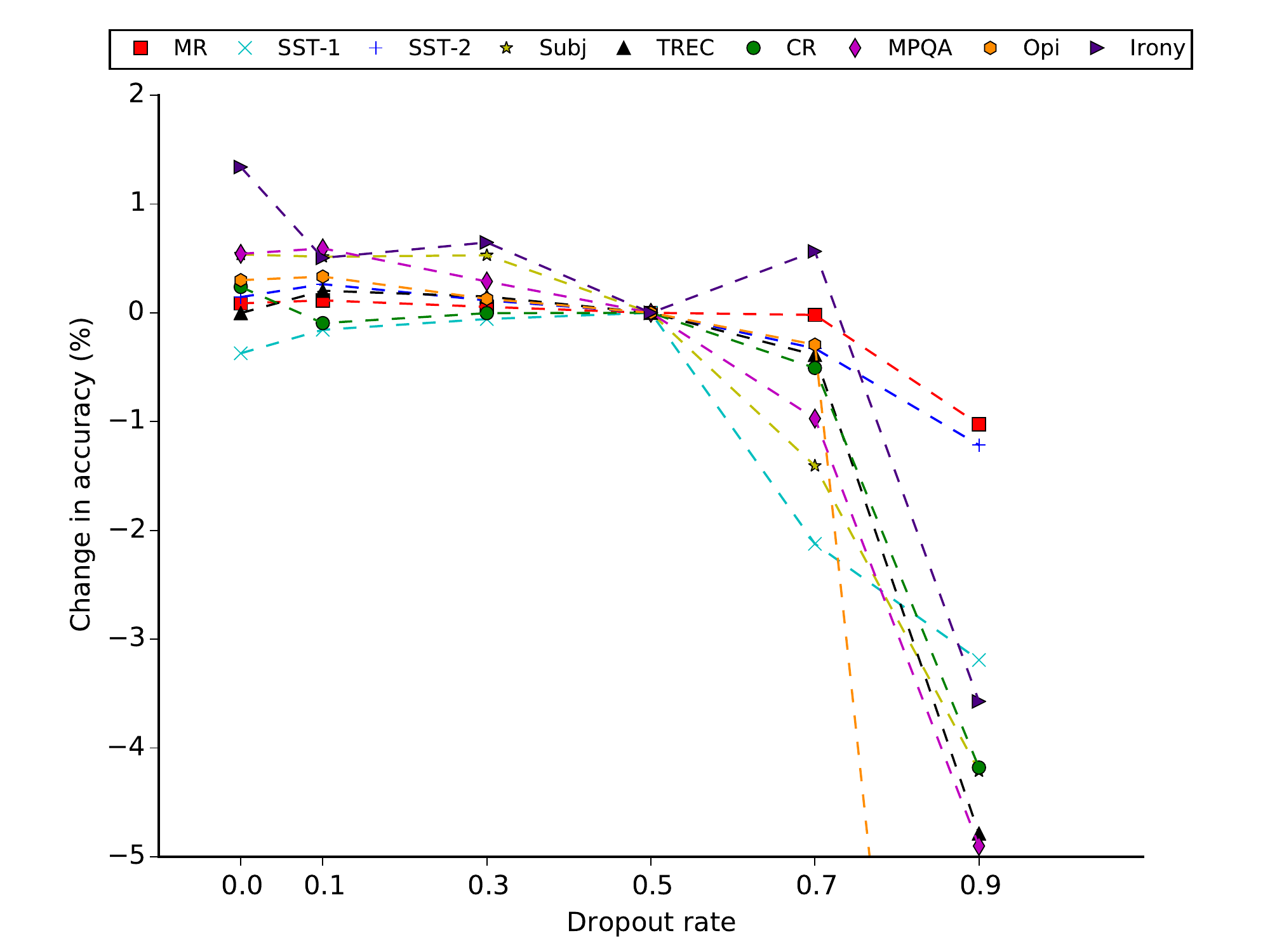}
\caption{Effect of dropout rate on the convolution layer (The accuracy when the dropout rate is 0.9 on the Opi dataset is not visible on the figure at this point, as in Fig. \ref{fig:nonstatic_dropout})}
\label{fig:dropout_CNN}
\end{figure}

We also experimented with applying dropout only to the convolution layer, but still setting the max norm constraint on the classification layer to 3, keeping all other settings exactly the same. This means we randomly set elements of the sentence matrix to 0 during training with probability $p$, and then multiplied $p$ with the sentence matrix at test time. The effect of dropout rate on the convolution layer is shown in Fig. \ref{fig:dropout_CNN}. Again we see that dropout on the convolution layer helps little, and large dropout rate dramatically hurts performance.

To summarize, contrary to some of the existing literature e \cite{srivastava2014dropout}, we found that dropout had little beneficial effect on CNN performance. We attribute this observation to the fact that one-layer CNN has a smaller number parameters than multi-layer deep learning models. Another possible explanation is that using word embeddings helps to prevent overfitting (compared to bag of words based encodings). However, we are not advocating completely foregoing regularization. Practically, we suggest setting the dropout rate to a small value (0.0-0.5) and using a relatively large max norm constraint, while increasing the number of feature maps to see whether more features might help. When further increasing the number of feature maps seems to degrade performance, it is probably worth increasing the dropout rate.



\section{Conclusions}
\label{section:conclusions}

We have conducted an extensive experimental analysis of CNNs for sentence classification. We conclude here by summarizing our main findings and deriving from these practical guidance for researchers and practitioners looking to use and deploy CNNs in real-world sentence classification scenarios.


\subsection{Summary of Main Empirical Findings}

\begin{itemize}
\item Prior work has tended to report only the mean performance on datasets achieved by models. But this overlooks variance due solely to the stochastic inference procedure used. This can be substantial: holding everything constant (including the folds), so that variance is due exclusively to the stochastic inference procedure, we find that mean accuracy (calculated via 10 fold cross-validation) has a range of up to 1.5 points. And the range over the AUC achieved on the irony dataset is even greater -- up to 3.4 points (see Table \ref{table:basic}). 
More replication should be performed in future work, and ranges/variances should be reported, to prevent potentially spurious conclusions regarding relative model performance. 

\item We find that, even when tuning them to the task at hand, the choice of input word vector representation (e.g., between word2vec and GloVe) has an impact on performance, however different representations perform better for different tasks. 
At least for sentence classification, both seem to perform better than using one-hot vectors directly. We note, however, that: (1) this may not be the case if one has a sufficiently large amount of training data, and, (2) the recent semi-supervised CNN model proposed by Johnson and Zhang \cite{johnson2015semi} may improve performance, as compared to the simpler version of the model considered here (i.e., proposed in \cite{johnson2014effective}).

\item The filter region size can have a large effect on performance, and should be tuned.
\item The number of feature maps can also play an important role in the performance, and increasing the number of feature maps will increase the training time of the model. 
\item 1-max pooling uniformly outperforms other pooling strategies. 
\item Regularization has relatively little effect on the performance of the model.

\end{itemize}

\subsection{Specific advice to practitioners}
\label{section:guidance}


Drawing upon our empirical results, we provide the following guidance regarding CNN architecture and hyperparameters for practitioners looking to deploy CNNs for sentence classification tasks. 

\begin{itemize}
  \item Consider starting with the basic configuration described in Table \ref{table:original} and using non-static word2vec or GloVe rather than one-hot vectors. 
  However, if the training dataset size is very large, it may be worthwhile to explore using one-hot vectors. Alternatively, if one has access to a large set of unlabeled in-domain data, \cite{johnson2015semi} might also be an option. 
  
  \item Line-search over the single filter region size to find the `best' single region size. A reasonable range might be 1$\sim$10. However, for datasets with very long sentences like CR, it may be worth exploring larger filter region sizes. Once this `best' region size is identified, it may be worth exploring combining multiple filters using regions sizes near this single best size, given that empirically multiple `good' region sizes always outperformed using only the single best region size.
   
   \item Alter the number of feature maps for each filter region size from 100 to 600, and when this is being explored, use a small dropout rate (0.0-0.5) and a large max norm constraint. Note that increasing the number of feature maps will increase the running time, so there is a trade-off to consider. Also pay attention whether the best value found is near the border of the range~\cite{bengio2012practical}. If the best value is near 600, it may be worth trying larger values.  
   
   \item Consider different activation functions if possible: ReLU and tanh are the best overall candidates. And it might also be worth trying no activation function at all for our one-layer CNN. 
   
   \item Use 1-max pooling; it does not seem necessary to expend resources evaluating alternative strategies.
   
   \item Regarding regularization: 
   When increasing the number of feature maps begins to reduce performance, try imposing stronger regularization, e.g., a dropout out rate larger than 0.5. 
   
   
   
   
   \item When assessing the performance of a model (or a particular configuration thereof), it is imperative to consider variance. Therefore, replications of the cross-fold validation procedure should be performed and variances and ranges should be considered. 
   
\end{itemize}

Of course, the above suggestions are applicable only to datasets comprising sentences with similar properties to the those considered in this work. And there may be examples that run counter to our findings here. Nonetheless, we believe these suggestions are likely to provide a reasonable starting point for researchers or practitioners looking to apply a simple one-layer CNN to real world sentence classification tasks. We emphasize that we selected this simple one-layer CNN in light of observed strong empirical performance, which positions it as a new standard baseline model akin to bag-of-words SVM and logistic regression. This approach should thus be considered prior to implementation of more sophisticated models.

We have attempted here to provide practical, empirically informed guidance to help data science practitioners find the best configuration for this simple model. We recognize that manual and grid search over hyperparameters is sub-optimal, and note that our suggestions here may also inform hyperparameter ranges to explore in random search or Bayesian optimization frameworks. 

\section{Acknowledgments}

This work was supported in part by the Army Research Office (grant W911NF-14-1-0442) and by The Foundation for Science and Technology, Portugal (grant UTAP-EXPL/EEIESS/0031/2014). This work was also made possible by the support of the Texas Advanced Computer Center (TACC) at UT Austin.

We thank Tong Zhang and Rie Johnson for helpful feedback. 

%
\bibliographystyle{acl.bst}
\bibliography{sigproc}  
%
%
\clearpage
\section*{Appendix}

\begin{table}[h]
\begin{center}
\begin{tabular}{c|c|c}

\bf Dataset & \bf Average length & \bf Maximum length \\ \hline
MR & 20 & 56 \\ \hline
SST-1 & 18 & 53 \\  \hline
SST-2 & 19 & 53\\ \hline
Subj & 23 & 120 \\ \hline
TREC & 10 & 37\\ \hline
CR & 19 & 105  \\ \hline
MPQA & 3 &36 \\ \hline
\end{tabular}
\end{center}
\caption{Average length and maximum length of the 7 datasets}
\label{table:length}
\end{table}

\begin{table}[h]
\footnotesize
\begin{center}
\begin{tabular}{c|c|c|c}
\hline \bf Dataset & \bf \bf {bow-LG} & \bf{wv-LG} & \bf {bow+wv-LG}  \\ \hline
MR & 78.24 &77.65 & 79.68  \\ \hline
SST-1 & 40.91 & 43.60 & 43.09 \\ \hline
SST-2 & 81.06 & 81.30 & 83.23 \\ \hline
Subj & 89.00 & 90.88 & 91.84\\ \hline
TREC &87.93 & 77.42 & 89.23 \\ \hline
CR & 77.59 & 80.79 & 80.39 \\ \hline
MPQA & 83.60 & 88.30 & 89.14\\ \hline
\end{tabular}
\end{center}
\caption{Performance of logistic regression}
\label{table:LG}
\end{table}

\begin{table*}
\tiny
\begin{center}
\begin{tabular}{c|c|c|c|c|c|c|c}

\bf Region size & \bf MR & \bf SST-1 & \bf SST-2 & \bf Subj & \bf TREC & \bf CR & \bf MPQA  \\ \hline 
 1 & 77.85 (77.47,77.97)& 44.91 (44.42,45.37) & 82.59(82.20,82.80)& 91.23 (90.96,91.48)& 85.82 (85.41,86.12) & 80.15 (79.27,80.89) & 88.53 (88.29,88.86) \\ \hline
 3 & 80.48 (80.26,80.65)& 46.64 (46.21,47.07) & 84.74 (84.47,85.00) & 92.71 (92.52,92.93)& 91.21 (90.88,91.52)& 81.01 (80.64,81.53) & 89.04 (88.71,89.27)\\ \hline
 5 & 81.13 (80.96,81.32)& 47.02 (46.74,47.40) & 85.31 (85.04,85.71) &92.89 (92.64,93.07)& 91.20 (90.96,91.43) & 81.78 (80.75,82.52) & 89.20 (88.99,89.37)\\ \hline
 7 & 81.65 (81.45,81.85) & 46.98 (46.70,47.37) & 85.57 (85.16,85.90) &92.95 (92.72,93.19)& 90.77 (90.53,91.15) & 82.16 (81.70,82.87)& 89.32 (89.17,89.41) \\ \hline
 10 & 81.43 (81.28,81.75) & 46.90 (46.50,47.56) & 85.60 (85.33,85.90) &92.90 (92.71,93.10)& 90.29 (89.89,90.52) & 82.53 (81.58,82.92) & 89.23 (89.03,89.52)\\ \hline
 15 & 81.26 (81.01,81.43) & 46.66 (46.13,47.23)& 85.33 (84.96,85.74) &92.82 (92.61,92.98)& 90.05 (89.68,90.28) & 82.49 (81.61,83.06) & 89.25 (89.03,89.44)\\ \hline
 20 & 81.06 (80.87,81.30) & 46.20 (45.40,46.72) & 85.02 (84.94,85.24) &92.72(92.47,92.87) & 90.01 (89.84,90.50) & 82.62 (82.16,83.03) & 89.16 (88.92,89.28)\\ \hline
 25 & 80.91 (80.73,81.10)& 46.17 (45.20,46.92) & 84.91 (84.49,85.39) &92.75 (92.45,92.96)& 89.99 (89.66,90.40) & 82.87 (82.21,83.45) & 89.16 (88.99,89.45)\\ \hline
 30 & 80.91 (80.72,81.05) & 46.02 (45.21,46.54) & 84.94 (84.63,85.25) &92.70 (92.50,92.90)& 89.90 (89.58,90.13) & 83.01 (82.44,83.38) & 89.15 (88.93,89.41)\\ \hline
\end{tabular}
\end{center}
\caption{Effect of single filter region size using non-static CNN.}
\label{table:filter_size}
\end{table*}

\begin{table*}
\tiny
\begin{center}
\begin{tabular}{c|c|c|c|c|c|c|c}

& \bf MR & \bf SST-1 & \bf SST-2 & \bf Subj & \bf TREC & \bf CR & \bf MPQA  \\ \hline 
1 &79.22 (79.02,79.57)& 45.46 (44.88,45.96) & 83.24 (82.93,83.67) &91.97 (91.64,92.17)& 85.86 (85.54,86.13) & 80.24 (79.64,80.62) & 88.25 (88.04,88.63)\\ \hline
3 &80.27 (79.94,80.51)& 46.18 (45.74,46.52) & 84.37 (83.96,94.70) &92.83 (92.58,93.06)& 90.33 (90.05,90.62) & 80.71 (79.72,81.37) & 89.37 (89.25,89.67)\\ \hline
5 &80.35 (80.05,80.65)& 46.18 (45.69,46.63) & 84.38 (84.04,84.61) &92.54 (92.44,92.68)& 90.06 (89.84,90.26) & 81.11 (80.54,81.55) & 89.50 (89.33,89.65)\\ \hline
7 &80.25 (79.89,80.60)& 45.96 (45.44,46.55) & 84.24 (83.40,84.59) &92.50 (92.33,92.68)& 89.44 (89.07,89.84) & 81.53 (81.09,82.05) & 89.44 (89.26,89.68)\\ \hline
10 &80.02 (79.68,80.17)& 45.65 (45.08,46.09) & 83.90 (83.40,84.37) &92.31 (92.19,92.50)& 88.81(88.53,89.03) & 81.19 (80.89,81.61) & 89.26 (88.96,89.60)\\ \hline
15 &79.59 (79.36,79.75)& 45.19 (44.67,45.62) & 83.64 (83.32,83.95) &92.02 (91.86,92.23)& 88.41 (87.96,88.71) & 81.36 (80.72,82.04) & 89.27 (89.04,89.49)\\ \hline
20 &79.33 (78.76,79.75)& 45.02 (44.15,45.77) & 83.30 (83.03,83.60) &91.87 (91.70,91.99)& 88.46 (88.21,88.85) & 81.42 (81.03,81.90) & 89.28 (88.90, 89.42)\\ \hline
25  &79.05 (78.91,79.21)& 44.61 (44.05,45.53) & 83.24 (82.82,83.70)&91.95 (91.59,92.16)& 88.23 (87.57,88.56)  & 81.16 (80.69,81.57)& 89.24 (88.87,89.42)\\ \hline
30 &79.04 (78.86,79.30)& 44.66 (44.42,44.91) & 83.09 (82.61,83.42) &91.85 (91.74,92.00)& 88.41 (87.98,88.67) & 81.28 (80.96,81.55)& 89.13 (88.91,89.33)\\ \hline
\end{tabular}
\end{center}
\caption{Effect of single filter region size using static CNN.}
\label{table:filter_static}
\end{table*}

\begin{table*}
\tiny
\begin{adjustwidth}{-0.8cm}{}
\centering
\begin{tabular}{c|c|c|c|c|c|c|c|c}
& \bf 10 & \bf 50 & \bf 100 & \bf 200 & \bf 400  & \bf 600 & \bf 1000 & \bf 2000  \\ \hline

MR &80.47 (80.14,80.99)& 81.25 (80.90,81.56) &81.17 (81.00,81.38) & 81.31 (81.00,81.60) & 81.41 (81.21,81.61)& 81.38 (81.09, 81.68) & 81.30 (81.15,81.39) & 81.40 (81.13,81.61)\\ \hline

SST-1&45.90 (45.14,46.41)&47.06 (46.58,47.59) &47.09 (46.50,47.66)& 47.09 (46.34,47.50) & 46.87 (46.41,47.43) & 46.84 (46.29,47.47) & 46.58 (46.26,47.14) & 46.75 (45.87,47.67) \\ \hline

SST-2 &84.26 (83.93,84.73) &85.23 (84.86,85.57) & 85.50 (85.31,85.66) &85.53 (85.24,85.69) & 85.56 (85.27,85.79) & 85.70 (85.57,85.93) & 85.75 (85.53,86.01) & 85.74 (85.49,86.02) \\ \hline

Subj &92.24 (91.74,92.43)&93.07 (92.94,93.28)&93.19 (93.08,93.45) & 93.29 (93.07,93.38) & 93.24 (92.96,93.39) & 93.34 (93.22,93.44) &93.32 (93.17,93.49) & 93.34 (93.05,93.49) \\ \hline 

TREC &90.64 (90.19,90.86) & 91.40 (91.12,91.59) &91.54 (91.17,91.90)& 91.54 (91.23,91.71) & 91.52 (91.30,91.70) & 91.50 (91.23,91.71) & 91.44 (91.26,91.56) & 91.54 (91.28,91.75) \\ \hline

CR &79.95 (79.36,80.41) &83.19 (82.32,83.50)&83.86 (83.52,84.15) & 84.30 (83.80,84.64) & 84.44 (84.14,85.02) & 84.62 (84.31,84.94) & 84.58 (84.35,84.85) & 84.47 (83.84,85.03)\\ \hline

MPQA &89.02 (88.89,89.19) &89.21 (88.97,89.41) & 89.21 (88.90,89.51) & 89.50 (89.27,89.68)& 89.57 (89.13,89.81) & 89.66 (89.35,89.90) & 89.55 (89.22,89.73) & 89.66 (89.47,89.94) \\ \hline

\end{tabular}
\caption{Performance of number of feature maps for each filter using non-static word2vec-CNN}
\label{table:feature_maps}
\end{adjustwidth}
\end{table*}

\begin{table*}
\tiny
\centering
\begin{tabular}{c|c|c|c|c|c|c|c|c}
& \bf 10 & \bf 50 & \bf 100 & \bf 200 & \bf 400  & \bf 600 & \bf 1000 & \bf 2000  \\ \hline

MR & 79.38 (78.88, 79.82) & 80.49 (80.16, 80.87) & 80.60 (80.27,80.85) & 80.76 (80.48,81.00) & 80.80 (80.56,81.11) & 80.79 (80.68,80.86) & 80.90 (80.67,81.16) & 80.84 (80.38,81.27)\\ \hline

SST-1 & 45.62 (45.28,46.01) & 46.33 (46.00,46.69) & 46.21 (45.68,46.85) & 46.23 (45.70, 46.99) & 46.10 (45.71,46.59) & 46.20 (45.85,46.55) & 46.56 (46.26,46.92) & 45.93 (45.57,46.27) \\ \hline

SST-2 & 83.38 (82.65,83.68) & 84.71 (84.46,85.27) & 84.89 (84.56,85.16) & 84.92 (84.81,85.18) & 84.98 (84.66,85.18) & 84.99 (84.29,85.44) & 84.90 (84.66,85.05) & 84.97 (84.79,85.14)\\ \hline

Subj & 91.84 (91.30,92.02) & 92.75 (92.61,92.88) & 92.89 (92.66,93.06) & 92.88 (92.75,92.97) & 92.91 (92.75,93.01) & 92.88 (92.75,93.03) & 92.89 (92.74,93.05) & 92.89 (92.64,93.11) \\ \hline 

TREC & 89.02 (88.62,89.31) & 90.51 (90.26, 90.82) & 90.62 (90.09,90.82) & 90.73 (90.48,90.99) & 90.72 (90.43,90.89) & 90.70 (90.51,91.03) & 90.71 (90.46,90.94) & 90.70 (90.53,90.87) \\ \hline

CR & 79.40 (78.76,80.03) & 82.57 (82.05,83.31) & 83.48 (82.99,84.06) & 83.83 (83.51,84.26) & 83.95 (83.36,84.60) & 83.96 (83.49, 84.47) & 83.95 (83.40,84.44) & 83.81 (83.30,84.28) \\ \hline

MPQA & 89.28 (89.04,89.45) & 89.53 (89.31,89.72) & 89.55 (89.18,89.81) & 89.73 (89.62,89.85) & 89.80 (89.65,89.96) & 89.84 (89.74,90.02) & 89.72 (89.57,89.88) & 89.82 (89.52,89.97) \\ \hline

\end{tabular}
\caption{Effect of number of feature maps for each filter using static word2vec-CNN}
\label{table:feature_maps_static}
\end{table*}

\begin{table}[h]
\footnotesize
\begin{center}
\begin{tabular}{c|c}
\hline \bf Dataset & \bf{One-hot vector CNN}   \\ \hline
MR & 77.83 (76.56,78.45) \\ \hline
SST-1 &  41.96 (40.29,43.46)\\ \hline
SST-2 & 79.80 (78.53,80.52) \\ \hline
Subj & 91.14 (90.38,91.53) \\ \hline
TREC & 88.28 (87.34,89.30) \\ \hline
CR & 78.22 (76.67,80.00) \\ \hline
MPQA & 83.94 (82.94,84.31) \\ \hline
\end{tabular}
\end{center}
\caption{Performance of one-hot vector CNN}
\label{table:one-hot}
\end{table}

\begin{table*}[t]
\tiny
\begin{adjustwidth}{}{}
\centering
\begin{tabular}{c|c|c|c|c|c|c|c}
 & \bf Sigmoid & \bf tanh &\bf Softplus &\bf Iden &\bf ReLU & \bf Cube & \bf {tahn-cube}\\ \hline
MR & 80.51 (80.22, 80.77) & 81.28 (81.07, 81.52) & 80.58 (80.17, 81.12) & 81.30 (81.09, 81.52) & 81.16 (80.81, 83.38) & 80.39 (79.94,80.83) & 81.22 (80.93,81.48)\\\hline

SST-1 & 45.83 (45.44, 46.31) & 47.02 (46.31, 47.73) & 46.95 (46.43, 47.45) & 46.73 (46.24,47.18) & 47.13 (46.39, 47.56) & 45.80 (45.27,46.51) & 46.85 (46.13,47.46)\\ \hline

SST-2 & 84.51 (84.36, 84.63) & 85.43 (85.10, 85.85) & 84.61 (84.19, 84.94) & 85.26 (85.11, 85.45) & 85.31 (85.93, 85.66) & 85.28 (85.15,85.55) & 85.24 (84.98,85.51) \\ \hline

Subj & 92.00 (91.87, 92.22) & 93.15 (92.93, 93.34) & 92.43 (92.21, 92.61) & 93.11 (92.92, 93.22) & 93.13 (92.93, 93.23) & 93.01 (93.21,93.43) & 92.91 (93.13,93.29)\\ \hline

TREC & 89.64 (89.38, 89.94) & 91.18 (90.91, 91.47)& 91.05 (90.82, 91.29) & 91.11 (90.82, 91.34) & 91.54 (91.17, 91.84) & 90.98 (90.58,91.47) & 91.34 (90.97,91.73)\\ \hline

CR & 82.60 (81.77, 83.05) & 84.28 (83.90, 85.11) & 83.67 (83.16, 84.26)& 84.55 (84.21, 84.69)& 83.83 (83.18, 84.21) & 84.16 (84.47,84.88) & 83.89 (84.34,84.89)\\ \hline

MPQA & 89.56 (89.43, 89.78) & 89.48 (89.16, 89.84) & 89.62 (89.45, 89.77) & 89.57 (89.31, 89.88) & 89.35 (88.88, 89.58) & 88.66 (88.55,88.77) & 89.45 (89.27,89.62)\\ 
\hline
\end{tabular}
\caption{Performance of different activation functions using non-static word2vec-CNN}
\label{table:activation}
\end{adjustwidth}
\end{table*}

\begin{table*}[!h]
\begin{adjustwidth}{}{}
\centering
\footnotesize
\begin{tabular}{c|c|c|c|c|c}
 & \bf Sigmoid &\bf tanh &\bf Softplus &\bf Iden &\bf ReLU \\ \hline
MR & 79.23 (79.11, 79.36) & 80.73 (80.29, 81.04) & 80.05 (79.76, 80.37) & 80.63 (80.26, 81.04) & 80.65 (80.44, 81.00)\\ \hline
TREC & 85.81 (85.65, 85.99) & 90.25 (89.92, 90.44) & 89.50 (89.36, 89.97)& 90.36 (90.23, 90.45) & 90.23 (89.85, 90.63)\\ \hline
CR & 81.14 (80.57, 82.01) & 83.51 (82.91,83.95) & 83.28 (82.67, 83.88)& 83.82 (83.50, 84.15) & 83.51 (82.54, 83.85)\\ \hline
SST-1 & 45.25 (44.65, 45.86)& 45.98 (45.68, 46.44) & 46.76 (46.41, 47.45)& 46.01 (45.60, 46.32) & 46.25 (45.70, 46.98)\\ \hline
SST-2 & 83.07 (82.48, 83.54)& 84.65 (84.36, 85.00) & 84.01 (83.57, 84.40) & 84.71 (84.40, 85.07) & 84.70 (84.31, 85.20)\\ \hline
Subj & 91.56 (91.39, 91.71) & 92.75 (92.60, 92.95) & 92.20 (92.08, 92.32) & 92.71 (92.51, 92.89) & 92.83 (92.67, 92.95)\\ \hline
MPQA & 89.43 (89.27, 89.56) & 89.75 (89.64, 89.86) & 89.45 (89.30, 89.56) & 89.75 (89.56, 89.87) & 89.66 (89.44, 90.00)\\ \hline
\end{tabular}
\caption{Performance of different activation function using static word2vec-CNN}
\label{table:activation_static}
\end{adjustwidth}
\end{table*}

\begin{table}[h]
\footnotesize
\centering
\begin{tabular}{c|c|c}
\bf {Pooling region} & \bf CR & \bf TREC \\ \hline
3 & 81.01 (80.73,81.28)& 88.89 (88.67,88.97)\\ \hline
10 & 80.74 (80.36,81.09) & 88.10 (87.82,88.47)\\ \hline
20 & 80.69 (79.72,81.32) & 86.45 (85.65,86.42)\\ \hline
30 & 81.13 (80.16,81.76)& 84.95 (84.65,85.14)\\ \hline
all & 80.17 (79.97,80.84) &83.30 (83.11,83.57)\\ \hline
\end{tabular} 
\caption{Performance of local average pooling region size using non-static word2vec-CNN (`all' means average pooling over the whole feature map resulting in one number)}
\label{table:average_pooling}
\end{table}

\begin{table*}[t]
\footnotesize
\centering
\begin{tabular}{c|c|c|c|c|c}
& \bf 1 (1-max) & \bf 5 & \bf 10 & \bf 15 & \bf 20 \\ \hline
MR & 81.25 (81.00,81.47) & 80.83 (80.69,80.91) & 80.05 (79.69,80.41) & 80.11 (79.89,80.36) & 80.05 (79.72,80.25) \\ \hline
SST-1 & 47.24 (46.90,47.65) & 46.63 (46.31,47.12) & 46.04 (45.27,46.61) & 45.91 (45.16,46.49) & 45.31 (44.90,45.63) \\ \hline
SST-2 & 85.53 (85.26,85.80) & 84.61(84.47,84.90) & 84.09 (83.94,84.22) & 84.02 (83.57,84.28) & 84.04 (83.74,84.34) \\ \hline

Subj & 93.18 (93.09,93.31)& 92.49 (92.33,92.61) & 92.66 (92.50,92.79) & 92.52 (92.33,92.96) & 92.58 (92.50,92.83) \\ \hline
TREC & 91.53 (91.26,91.78) & 89.93 (89.75,90.09) & 89.73 (89.61,89.83) & 89.49(89.31,89.65) & 89.05(88.85,89.34) \\ \hline
CR& 83.81 (83.44,84.37) & 82.70 (82.14,83.11) & 82.46 (82.17,82.76) & 82.26 (81.86, 82.90) & 82.09 (81.74,82.34) \\ \hline
MPQA& 89.39 (89.14, 89.58)  & 89.36 (89.17,89.57) & 89.14 (89.00,89.45) & 89.31 (89.18,89.48) & 88.93 (88.82,89.06) \\ \hline
\end{tabular} 
\caption{Performance of global k-max pooling using non-static word2vec-CNN}
\end{table*}

\begin{table*}[t]
\footnotesize
\centering
\begin{tabular}{c|c|c|c|c|c}
 & \bf max,3 & \bf max,10 & \bf max,20 & \bf max,30 & \bf max,all (1-max)\\ \hline
 
 MR & 79.75 (79.47,80.03) & 80.20 (80.02,80.35) & 80.68 (80.14,81.21) & 80.99 (80.65,81.30) & 81.28 (81.16,81.54)\\ \hline
 
SST-1 & 44.98 (44.06,45.68) & 46.10(45.37,46.84) & 46.75 (46.35,47.36) & 47.02 (46.59,47.59) & 47.00 (46.54,47.26) \\ \hline
SST-2 & 83.69(83.46,84.07) & 84.63 (84.44,84.88) & 85.18 (84.64,85.59) & 85.38 (85.31,85.49) & 85.50 (85.31,85.83) \\ \hline

Subj & 92.60 (92.28,92.76) & 92.87 (92.69,93.17) & 93.06 (92.81,93.19) & 93.13 (92.79,93.32) & 93.20 (93.00,93.36) \\ \hline

TREC & 90.29 (89.93,90.61) & 91.42 (91.16,91.71) & 91.52 (91.23,91.72) & 91.47 (91.15,91.64) & 91.56 (91.67,91.88)\\ \hline

CR & 81.72 (81.21,82.20) & 82.71 (82.06,83.30) & 83.44(83.06,83.90) & 83.70 (83.31,84.25) & 83.93 (83.48,84.39) \\ \hline
MPQA & 89.15 (88.83,89.47) & 89.39 (89.14,89.56) & 89.30 (89.16,89.60) & 89.37 (88.99,89.61) & 89.39 (89.04,89.73)\\ \hline
\end{tabular} 
\caption{Performance of local max pooling using non-static word2vec-CNN}
\label{table:pooling}
\end{table*}

\begin{table*}
\tiny
\centering
\begin{tabular}{c|c|c|c|c|c|c|c}

&\bf None & \bf 0.0 & \bf 0.1 & \bf 0.3 & \bf 0.5 &  \bf 0.7 & \bf 0.9  \\ \hline
 MR &81.15 (80.95,81.34) & 81.24 (80.82, 81.63 )&81.22 (80.97 ,81.61 )&81.30 (81.03 ,81.48 )&81.33 (81.02, 81.74 )&81.16 (80.83, 81.57 )&80.70 (80.36, 80.89) \\ \hline
 
 SST-1 &46.30 (45.81,47.09) & 45.84 (45.13 ,46.43 )&46.10 (45.68, 46.36 )&46.61 (46.13, 47.04 )&47.09 (46.32, 47.66 )&47.19 (46.88 ,47.46 )&45.85 (45.50, 46.42 )\\ \hline
 
 SST-2 &85.42 (85.13,85.23) & 85.53 (85.12 ,85.88 )&85.69 (85.32, 86.06 )&85.58 (85.30, 85.76 )&85.62 (85.25, 85.92 )&85.41 (85.18, 85.65 )&84.49 (84.35, 84.82 ) \\ \hline
 
 Subj &93.23 (93.09,93.37)& 93.21 (93.09 ,93.31 )&93.27 (93.12 ,93.45 )&93.28 (93.06, 93.39 )&93.14 (93.01, 93.32 )&92.94 (92.77 ,93.08 )&92.03 (91.80 ,92.24 ) \\ \hline
 
 TREC &91.38 (91.18,91.59)& 91.39 (91.13 ,91.66 )&91.41 (91.26, 91.63 )&91.50 (91.22 ,91.76 )&91.54 (91.41, 91.68 )&91.45 (91.17, 91.77 )&88.83 (88.53 ,89.19 ) \\ \hline
 
CR &84.36 (84.06,84.70)& 84.04 (82.91, 84.84 )&84.22 (83.47, 84.60 )&84.09 (83.72, 84.51 )&83.92 (83.12, 84.34 )&83.42 (82.87, 83.97 )&80.78 (80.35, 81.34 ) \\ \hline

MPQA &89.30 (88.91,89.68)& 89.30 (89.01, 89.56 )&89.41 (89.19, 89.64 )&89.40 (89.18, 89.77 )&89.25 (88.96, 89.60 )&89.24 (88.98, 89.50 )&89.06 (88.93, 89.26 ) \\ 
\hline
\end{tabular}

\caption{Effect of dropout rate using non-static word2vec-CNN}
\label{table:dropout_nonstatic}
\end{table*}

\begin{table*}
\tiny
\centering
\begin{tabular}{c|c|c|c|c|c|c|c}

& \bf None &  \bf 0.0 & \bf 0.1 & \bf 0.3 & \bf 0.5 &  \bf 0.7 & \bf 0.9  \\ \hline
 MR &80.19(79.95,80.39)& 80.37 (80.03, 80.66 )&80.54 (80.13, 80.90 )&80.46 (80.20, 80.63 )&80.66 (80.34, 81.10 )&80.70 (80.31, 80.95 )&79.88 (79.57, 80.06 ) \\ \hline
 
 SST-1 &45.11 (44.57,45.64)& 45.40 (45.00 ,45.72 )&45.08 (44.45, 45.70 )&45.94 (45.55, 46.45 )&46.41 (45.89, 46.92 )&46.87 (46.60 ,47.24 )&45.37 (45.18, 45.65 )\\ \hline
 
 SST-2 &84.58 (84.24,84.87)& 84.70 (84.34, 84.96 )&84.63 (84.41 ,84.95 )&84.80 (84.54, 84.99 )&84.95 (84.52, 85.29 )&84.82 (84.61 ,85.15 )&83.66 (83.45, 83.89 ) \\ \hline
 
 Subj & 92.88 (92.58,93.03) &92.82 (92.57 ,93.14 )&92.81 (92.71, 92.90 )&92.89 (92.64, 93.05 )&92.86 (92.77, 93.04 )&92.71 (92.51 ,92.93 )&91.60 (91.50, 91.79 )\\ \hline

TREC &90.55 (90.26,90.94)& 90.69 (90.36 ,90.93 )&90.84 (90.67, 91.06 )&90.75 (90.56, 90.95 )&90.71 (90.46, 91.10 )&89.99 (89.67,90.16 )&85.32 (85.01, 85.57 )\\ \hline

 CR &83.53 (82.96,84.15)& 83.46 (83.03 ,84.04 )&83.60 (83.22 ,83.87 )&83.63 (83.03, 84.08 )&83.38 (82.70, 83.67 )&83.32 (82.72 ,84.07 )&80.67 (80.12, 81.01 ) \\ \hline
 
 MPQA & 89.51 (89.42,89.67) & 89.36 (89.12 89.63 )&89.52 (89.32 89.68 )&89.55 (89.28 89.77 )&89.53 (89.37 89.79 )&89.52 (89.29 89.70 )&88.91 (88.76 89.12 )\\ \hline
\end{tabular}

\caption{Effect of dropout rate using static word2vec-CNN}
\label{table:dropout_static}
\end{table*}

\begin{table*}
\tiny
\centering
\begin{tabular}{c|c|c|c|c|c|c|c}

&\bf None & \bf 0.0 & \bf 0.1 & \bf 0.3 & \bf 0.5 &  \bf 0.7 & \bf 0.9  \\ \hline
 MR &  81.29 (81.05 81.55 )&81.48 (81.29 81.83 )&81.31 (81.09 81.62 )&81.50 (81.36 81.73 )&81.23 (80.91 81.41 )&81.21 (80.94 81.53 )&80.72 (80.47 80.95) \\ \hline
 SST-1 & 46.52 (46.32 46.75 )&46.25 (45.87 46.88 )&46.59 (46.21 47.14 )&46.58 (46.19 47.24 )&46.80 (46.31 47.43 )&47.41 (47.07 48.04 )&47.05 (46.50 47.44) \\ \hline
 SST-2 & 85.56 (85.20 86.05 )&85.82 (85.69 85.97 )&85.89 (85.63 86.00 )&85.85 (85.69 86.05 )&85.69 (85.61 85.86 )&85.52 (85.31 85.66 )&84.78 (84.58 84.95 ) \\ \hline
 Subj & 93.38 (93.17 93.48 )&93.29 (93.00 93.54 )&93.38 (93.20 93.46 )&93.37 (93.30 93.44 )&93.29 (93.23 93.37 )&93.13 (93.04 93.22 )&92.32 (92.22 92.45 ) \\ \hline
 TREC & 91.27 (91.17 91.49 )&91.53 (91.34 91.78 )&91.46 (91.40 91.52 )&91.63 (91.47 91.75 )&91.54 (91.42 91.74 )&91.27 (91.14 91.34 )&89.95 (89.80 90.26 ) \\ \hline
 CR & 84.87 (84.58 85.26 )&85.01 (84.63 85.49 )&84.72 (84.01 85.26 )&84.56 (84.28 84.79 )&84.42 (84.08 84.81 )&84.40 (84.08 84.65 )&82.69 (82.25 83.06 ) \\ \hline
 MPQA & 89.56 (89.31 89.71 )&89.52 (89.39 89.73 )&89.49 (89.27 89.83 )&89.59 (89.40 89.84 )&89.43 (89.16 89.54 )&89.62 (89.52 89.78 )&89.04 (88.92 89.15 ) \\ \hline
 
\end{tabular}

\caption{Effect of dropout rate when feature map is 500 using non-static word2vec-CNN}
\label{table:dropout_nonstatic_large}
\end{table*}
\label{table:kpooling}

\begin{table*}
\tiny
\centering
\begin{tabular}{c|c|c|c|c|c|c}

&\bf 0.0 & \bf 0.1 & \bf 0.3 & \bf 0.5 &  \bf 0.7 & \bf 0.9  \\ \hline
 MR & 81.16 (80.80 81.57 )&81.19 (80.98 81.46 )&81.13 (80.58 81.58 )&81.08 (81.01 81.13 )&81.06 (80.49 81.48 )&80.05 (79.92 80.37)\\ \hline
 SST-1 & 45.97 (45.65 46.43 )&46.19 (45.71 46.64 )&46.28 (45.83 46.93 )&46.34 (46.04 46.98 )&44.22 (43.87 44.78 )&43.15 (42.94 43.32) \\ \hline
 SST-2 & 85.50 (85.46 85.54 )&85.62 (85.56 85.72 )&85.47 (85.19 85.58 )&85.35 (85.06 85.52 )&85.02 (84.64 85.31 )&84.14 (83.86 84.51) \\ \hline
 Subj & 93.21 (93.13 93.31 )&93.19 (93.07 93.34 )&93.20 (93.03 93.39 )&92.67 (92.40 92.98 )&91.27 (91.16 91.43 )&88.46 (88.19 88.62) \\ \hline
 TREC & 91.41 (91.22 91.66 )&91.62 (91.51 91.70 )&91.56 (91.46 91.68 )&91.41 (91.01 91.64 )&91.03 (90.82 91.23 )&86.63 (86.15 86.90) \\ \hline
 CR & 84.21 (83.81 84.62 )&83.88 (83.54 84.11 )&83.97 (83.73 84.16 )&83.97 (83.75 84.18 )&83.47 (82.86 83.72 )&79.79 (78.89 80.38 )\\ \hline
 MPQA & 89.40 (89.15 89.56 )&89.45 (89.26 89.60 )&89.14 (89.08 89.20 )&88.86 (88.70 89.05 )&87.88 (87.71 88.18 )&83.96 (83.76 84.12) \\ \hline

 \end{tabular}

\caption{Effect of dropout rate on convolution layer using non-static word2vec-CNN}
\label{table:dropout_CNN}
\end{table*}

\begin{table*}[]
\centering
\tiny
\begin{tabular}{c|c|c|c|c|c|c|c}
 & \bf MR & \bf SST-1 & \bf SST-2 & \bf Subj & \bf TREC & \bf CR & \bf MPQA \\ \hline
1 & 81.02 (80.75 ,81.29) & 46.93 (46.57, 47.33) & 85.02 (84.76,85.22)& 92.49 (92.35 92.63) & 90.90 (90.62 91.20)&83.06 (82.50 83.42) &89.17 (88.97 89.36)\\ \hline
2 & 81.33 (81.04 ,81.71)& 47.11 (46.77, 47.43)& 85.40 (84.98,85.67)&92.93 (92.82 93.15)&91.44 (91.20 91.60)&84.00 (83.57 84.34)&89.31 (89.17 89.54)\\ \hline
3 &81.29 (80.96, 81.59)&47.29 (46.90 ,47.82)& 85.47 (85.17,85.77)&93.21 (93.03 93.37)&91.44 (91.18 91.68)&83.89 (83.24 84.47)&89.18 (88.84 89.40) \\ \hline
4 &81.38 (81.21, 81.68)&46.91 (46.22 ,47.38) & 85.33 (85.25,85.72)&93.08 (92.96 93.22)&91.56 (91.26 91.90)&84.00 (83.21 84.60)&89.27 (89.11 89.41)\\ \hline
5 &81.22 (81.03, 81.49)&46.93 (46.44 ,47.38) & 85.46 (84.98,85.73)&93.14 (92.90 93.33)&91.58 (91.39 91.87)&83.99 (83.73 84.31)&89.33 (89.02 89.55)\\ \hline
10 & 81.19 (80.94 ,81.42)&46.74 (46.19, 47.12) & 85.41 (85.04,85.83)&93.11 (92.99 93.32)&91.58 (91.29 91.81)&83.94 (83.04 84.61)&89.22 (89.01 89.40)\\ \hline
15 & 81.12 (80.87, 81.29)&46.91 (46.58 ,47.48) & 85.47 (85.23,85.74)&93.15 (92.99 93.29)&91.58 (91.37 91.84)&83.92 (83.40 84.54)&89.30 (88.93 89.66)\\ \hline
20 & 81.13 (80.64, 81.33)&46.96 (46.62 ,47.31) & 85.46 (85.17,85.64)&93.10 (92.98 93.19) &91.54 (91.28 91.73)&84.09 (83.59 84.53)&89.28 (88.92 89.43)\\ \hline
25 & 81.22 (80.82, 81.66)&47.02 (46.73, 47.67) & 85.42 (85.16,85.78)&93.09 (92.95 93.25)&91.45 (91.22 91.62)&83.91 (83.24 84.40)&89.33 (89.05 89.61)\\ \hline
30 & 81.19 (80.79 ,81.43)&46.98 (46.63 ,47.59) & 85.48 (85.27,85.79)&93.06 (92.84 93.43)&91.55 (91.26 91.84)&83.94 (83.02 84.35)&89.26 (89.10 89.54)\\ \hline
None & 80.19(79.95,80.39) & 46.30 (45.81,47.09) & 85.42 (85.13,85.23) & 93.23 (93.09,93.37) & 91.38 (91.18,91.59) & 84.36 (84.06,84.70) & 89.30 (88.91,89.68) \\ \hline
\end{tabular}
\caption{Effect of constraint on $l2$ norm using non-static word2vec-CNN}
\label{table:l2}
\end{table*}

\begin{table*}[!h]
\centering
\tiny
\begin{tabular}{c|c|c|c|c|c|c|c}
 & \bf MR & \bf SST-1 & \bf SST-2 & \bf Subj & \bf TREC & \bf CR & \bf MPQA \\ \hline
1 & 81.02 (80.75 ,81.29) & 46.93 (46.57, 47.33) & 85.02 (84.76,85.22)& 92.49 (92.35 92.63) & 90.90 (90.62 91.20)&83.06 (82.50 83.42) &89.17 (88.97 89.36)\\ \hline
2 & 81.33 (81.04 ,81.71)& 47.11 (46.77, 47.43)& 85.40 (84.98,85.67)&92.93 (92.82 93.15)&91.44 (91.20 91.60)&84.00 (83.57 84.34)&89.31 (89.17 89.54)\\ \hline
3 &81.29 (80.96, 81.59)&47.29 (46.90 ,47.82)& 85.47 (85.17,85.77)&93.21 (93.03 93.37)&91.44 (91.18 91.68)&83.89 (83.24 84.47)&89.18 (88.84 89.40) \\ \hline
4 &81.38 (81.21, 81.68)&46.91 (46.22 ,47.38) & 85.33 (85.25,85.72)&93.08 (92.96 93.22)&91.56 (91.26 91.90)&84.00 (83.21 84.60)&89.27 (89.11 89.41)\\ \hline
5 &81.22 (81.03, 81.49)&46.93 (46.44 ,47.38) & 85.46 (84.98,85.73)&93.14 (92.90 93.33)&91.58 (91.39 91.87)&83.99 (83.73 84.31)&89.33 (89.02 89.55)\\ \hline
10 & 81.19 (80.94 ,81.42)&46.74 (46.19, 47.12) & 85.41 (85.04,85.83)&93.11 (92.99 93.32)&91.58 (91.29 91.81)&83.94 (83.04 84.61)&89.22 (89.01 89.40)\\ \hline
15 & 81.12 (80.87, 81.29)&46.91 (46.58 ,47.48) & 85.47 (85.23,85.74)&93.15 (92.99 93.29)&91.58 (91.37 91.84)&83.92 (83.40 84.54)&89.30 (88.93 89.66)\\ \hline
20 & 81.13 (80.64, 81.33)&46.96 (46.62 ,47.31) & 85.46 (85.17,85.64)&93.10 (92.98 93.19) &91.54 (91.28 91.73)&84.09 (83.59 84.53)&89.28 (88.92 89.43)\\ \hline
25 & 81.22 (80.82, 81.66)&47.02 (46.73, 47.67) & 85.42 (85.16,85.78)&93.09 (92.95 93.25)&91.45 (91.22 91.62)&83.91 (83.24 84.40)&89.33 (89.05 89.61)\\ \hline
30 & 81.19 (80.79 ,81.43)&46.98 (46.63 ,47.59) & 85.48 (85.27,85.79)&93.06 (92.84 93.43)&91.55 (91.26 91.84)&83.94 (83.02 84.35)&89.26 (89.10 89.54)\\ \hline
None & 80.19(79.95,80.39) & 45.11 (44.57,45.64) & 84.58 (84.24,84.87) & 92.88 (92.58,93.03) & 90.55 (90.26,90.94) & 83.53 (82.96,84.15) & 89.51 (89.42,89.67) \\  \hline
\end{tabular}
\caption{Effect of constraint on $l2$-norms using static word2vec-CNN}
\label{table:l2_static}
\end{table*}

\end{document}